\documentclass[preprint,12pt]{elsarticle}

\makeatletter
\def\ps@pprintTitle{%
  \let\@oddhead\@empty
  \let\@evenhead\@empty
  \def\@oddfoot{\reset@font\hfil\thepage\hfil}
  \let\@evenfoot\@oddfoot
}
\makeatother

\usepackage{amsmath,amssymb}
\usepackage{float}
\usepackage{graphicx}
\usepackage{booktabs}
\usepackage{hyperref}
\usepackage{color}
\definecolor{Blue}{rgb}{0,0,1}

\definecolor{Orange}{rgb}{1,0.5,0}

\definecolor{Green}{rgb}{0,1,0}

\usepackage{amsmath,amssymb}
\usepackage{algorithm,algpseudocode}
\algnewcommand{\Inputs}[1]{%
  \State \textbf{Input:}
  \Statex \hspace*{\algorithmicindent}\parbox[t]{.8\linewidth}{\raggedright #1}
}
\algnewcommand{\Initialize}[1]{%
  \State \textbf{Initialize:}
  \Statex \hspace*{\algorithmicindent}\parbox[t]{.8\linewidth}{\raggedright #1}
}
\algnewcommand{\Outputs}[1]{%
  \State \textbf{Output:}
  \Statex \hspace*{\algorithmicindent}\parbox[t]{.8\linewidth}{\raggedright #1}
}

\usepackage{verbatim}
\usepackage{makecell}
\usepackage{subcaption}
\begin{document}

\begin{frontmatter}

\title{Categorizing Items with Short and Noisy Descriptions using Ensembled Transferred Embeddings}

\author[TAU]{Yonatan Hadar}
\ead{had.yonatan@gmail.com}
\author[TAU]{Erez Shmueli\corref{CORR}}
\ead{shmueli@tau.ac.il}

\cortext[CORR]{Corresponding author}
\address[TAU]{Department of Industrial Engineering, Tel-Aviv University, P.O. Box 39040, Tel Aviv 6997801, Israel}

\begin{abstract}
Item categorization is a machine learning task which aims at classifying e-commerce items, typically represented by textual attributes, to their most suitable category from a predefined set of categories.
An accurate item categorization system is essential for improving both the user experience and the operational processes of the company.
In this work, we focus on item categorization settings in which the textual attributes representing items are noisy and short, and labels (i.e., accurate classification of items into categories) are not available.
In order to cope with such settings, we propose a novel learning framework, Ensembled Transferred Embeddings (ETE), which relies on two key ideas:
1) labeling a relatively small sample of the target dataset, in a semi-automatic process, and
2) leveraging other datasets from related domains or related tasks that are large-scale and labeled, to extract ``transferable embeddings''.
Evaluation of ETE on a large-scale real-world dataset provided to us by PayPal, shows that it significantly outperforms traditional as well as state-of-the-art item categorization methods.
\end{abstract}

\begin{keyword}
Item-categorization; Product-categorization; Text-classification; Deep-neural-networks;
Embeddings;
Transfer-learning
\end{keyword}

\end{frontmatter}

\section{Introduction}
\label{intro}
Online shopping websites have become extremely popular in recent years.
In particular, marketplaces such as eBay.com and Amazon.com accept millions of new items every day \citep{cevahir2016large}.
To better cope with the enormous number of items, companies typically organize items into a predefined set of categories \citep{shen2012large}.
Such categorization of items\footnote{While we name this task ``item categorization'', it is important to note that a more accurate term would be ``item classification''. Nevertheless, we stick with the term ``item categorization'' as it is widely used in the literature, for example by \cite{shen2012large,ha2016large}.} is essential for enhancing user experience, where it allows users to search and navigate more easily between items, receive better recommendations for relevant items, and view customized descriptions of items.
Categorization of items is also important for improving operational processes of the company, such as targeted advertising, determining shipping and handling fees, fraud detection, identification of duplicate items, and enforcement of company policies (e.g., which items are allowed or forbidden to be sold on the website) \citep{shen2012large}.

While many item attributes can only be assigned by humans, the category of a given item can be inferred automatically from other attributes of that item. 
Automatic categorization of items is important both for saving the costs associated with manual labeling of millions of items every day and for ensuring the consistency of categories, which quickly becomes a serious problem when several individuals are involved in the labeling process \citep{kozareva2015everyone}. 

Indeed, many studies have investigated the problem of item categorization as a machine learning text classification task.
Earlier studies relied on representing text as a vector in a multidimensional space of identifiers (e.g., single index terms or N-grams) using various weighting schemes (e.g., TF-IDF) \citep{ding2002goldenbullet,yu2012product,mathivanan2018improving,sun2014chimera}.
However, these methods share the following limitations:
1) they typically produce training matrices which are very high dimensional and consequently very sparse;
2) the semantic meaning of words and their relationship to other words is typically ignored; and
3) the context of words (e.g., where they appear in the sentence, after which word, etc.) is rarely taken into account.

With the growing popularity of deep learning, studies have started to show the applicability of deep learning also in the domain of item categorization  \citep{das2016large,ha2016large,cevahir2016large,li2018don,krishnan2019large,chen2019fine}.
For example, \cite{ha2016large} demonstrated how using multiple Recurrent Neural Network (RNN) layers for encoding textual item attributes can significantly outperform traditional bag-of-words methods.
More recently, \cite{krishnan2019large} showed how to combine both textual and non-textual item attributes as part of Long Short Term Memory network (LSTM) and Convolutional Neural Network (CNN) layers for improving classification performance.
However, a major limitation of deep learning methods is the need for very large labeled datasets to train on.

Recent studies have tried to overcome the aforementioned limitation of deep learning models by using general purpose embeddings such as Word2Vec \citep{kozareva2015everyone} and general purpose pretrained models such as BERT \citep{zaheraprobert,Yang2020BertWD} and CamemBERT \citep{verma2018deep,leecbb2020CBB}.
In both cases, the idea is to train once deep learning models on a very large general-purpose dataset, usually based on news articles, books, or Wikipedia pages.
Then, the obtained embeddings or pretrained models are adjusted to the problem at hand, by using a considerably smaller labeled dataset from that problem's domain.
For example, the two best-performing solutions \citep{verma2018deep, leecbb2020CBB} in the e-commerce item categorization challenge as part of SIGIR 2020, used CamemBERT as a text encoder for item descriptions. 
However, as we demonstrate later in this paper, the general-purpose nature of these methods makes them less suitable for domain specific problems.

In this paper, we focus on item categorization settings in which:
1) item descriptions are relatively short and noisy, and
2) labeled data for the target dataset is unavailable.
Such settings entail that deep learning techniques cannot be applied directly on the target dataset, and general-purpose embedding might be less appropriate to use since the text distribution of the target dataset may differ greatly from that of the general-purpose corpus they were trained with.

To address such settings, we propose a novel learning framework, Ensembled Transferred Embedding (ETE), which has four main steps:
1) manually label a small sample dataset;
2) extract embeddings from related large-scale labeled datasets;
3) train transferred models using the extracted transferred embeddings and the labels of the sample dataset; and
4) build an ensemble to combine the outputs of the different transferred models into a single prediction.
We then show the applicability of the proposed framework to the item categorization task in settings for which item descriptions are noisy and short, and labels are not available.


Extensive evaluation that we conducted, using a large-scale real-world invoice dataset provided to us by PayPal, shows that our method significantly outperforms all other considered traditional (e.g., TF-IDF) as well as state-of-the-art (e.g., methods based on general purpose pretrained models such as BERT) item categorization methods.

\medskip
\noindent
The contribution of this paper is three-fold:
\begin{itemize}
\item We propose the ETE learning framework which relies on labeling a relatively small sample of the target dataset, in a semi-automatic process, and
leveraging other large-scale labeled datasets from related domains or related tasks.
\item We show the applicability of the proposed framework for the case of item categorization with short and noisy item descriptions.
\item An extensive evaluation that we conducted demonstrates the superiority of our method compared to traditional as well as state-of-the-art text classification methods.
\end{itemize}

\medskip
The rest of this paper is structured as follows:
In Section \ref{related_work}, we review the background and related work to this study.
In Section \ref{datasets}, we describe the datasets we used in this study.
In ection \ref{method}, we describe the proposed ETE learning framework and how it can be applied to our item categorization setting.
Section \ref{results} discusses the experimental setting and the results.
Section \ref{conclusions} summarizes this paper and suggests directions for future work.

\section{Related Work}\label{related_work}

In this section, we discuss two fields of research which are highly relevant to our study:
(1) classification tasks with limited labeled data and
(2) item categorization tasks.

\subsection{Classification tasks with limited labeled data}
Deep learning has become a very popular method for text classification in recent years, due to its ability to improve the accuracy of previous state-of-the-art methods on several benchmarks \citep{ruder2019neural}.
However, these improvements required hundreds of thousands to millions labeled training examples, which in many cases can be very time consuming and/or expensive to acquire. 
This challenge has contributed to the emergence and development of multiple active research fields addressing settings with limited labeled data  \citep{hedderich2020survey}.

\textbf{Distant \& Weak Supervision:}
Weak supervision is a simple strategy to train machine learning models using weak labels obtained by automated methods \citep{wang2019clinical}.
\cite{bach2019snorkel} showed how an unlabeled dataset can be labeled using a set of heuristics designed by domain experts.
For example in a product classification task, data can be labeled based on predefined keywords, a knowledge base of products names, or by mapping the output categories of an unsupervised topic model to a predefined set of categories.
Various labeling functions can later be combined using a generative model and used to assign probabilistic labels to training instances.
\cite{mekala2020contextualized} used only the class name and some related seed words generated by a domain expert in order to create a labeled dataset.
Then, the seed words were iteratively expanded and reduced using Pseudo-Labels and contextualized word embeddings.
\cite{jiang2011target} leveraged massive noisy-labeled tweets selected by positive and negative emoticons as a training set and built sentiment classifiers on these weak labels.
The setting considered in this paper assumes that an external knowledge base or a domain expert that can create labeling functions or class seed words are not available, and therefore distant \& weak supervision methods cannot be directly applied.

\textbf{Few shot learning:}
Few shot learning refers to a setting in which a classifier must be adapted to accommodate new classes which were not available during the training phase, given only a few examples of each of these classes \citep{snell2017prototypical}.
The main approaches to address this problem include:
1) augmenting the small labeled dataset,
2) projecting the data into a more suitable embedding space, and
3) performing weight initialization using a meta learning model \citep{wang2020generalizing}.
For example, \cite{vinyals2016matching} suggested a model that creates an embedding for a new image and compare it to the embedding of a small set of images from the training set.
The algorithm tries to minimize the cosine distance to images in the same class and maximize the distance to other classes.
In this way, when a new class appears, the algorithm can compare the learned embedding of the new class to images from other classes.
\cite{snell2017prototypical} suggested an improved model that form class prototypes from the average embedding of each class and compares the new image embedding to the prototypes and not to all training images.
\cite{finn2017model} suggested a different approach by which meta-learning is used to find a good weight initialization to the neural network.
In this way, only a small number of gradient steps are performed to finetune the network on the small training set of new classes, and overfitting is avoided.
\cite{antoniou2017data} trained a generative adversarial neural network (GAN) that creates new images in the same class of an input image.
In this way, the small training set of each class can be augmented with new syntactic images that were generated by the GAN.
The few shot learning setting is considerably different from the setting considered in this paper, in which the entire set of classes is known during the training phase, but labels are not available. 

\textbf{Transfer learning:}
Transfer learning is a method aimed to reduce the need for labeled target data by transferring learned representations from a related model trained on a large labeled dataset \citep{hedderich2020survey}.
For example, in computer vision, it is common to start the training of a new deep neural network with the weights of a neural network that was pre-trained on the large Imagenet dataset \citep{ruder2019neural}.
\cite{mikolov2013distributed} suggested Word2Vec, a method for creating unsupervised word embeddings by predicting the surrounding words in a sentence or a document.
Word2Vec vectors are commonly used to initialize the first layer of a neural network in many NLP models and showed great benefit in improving model accuracy.
\cite{howard2018universal} trained a deep LSTM neural network for a language model task on a large unlabeled dataset.
Then, the LSTM network was fine-tuned on a smaller target dataset.
This method improved the accuracy of the models drastically and showed that after the language model was pre-trained and finetuned on 100 labeled data instances, the accuracy of the model matched the accuracy of a model trained on the full dataset without pre-training.
\cite{devlin2018bert} introduced the BERT model, A Transformer network trained on very large corpus of unlabeled data on the task of masked language modeling and next sentence prediction.
BERT has shown state-of-the-art results when finetuned in several tasks such as natural language inference and question answering.
Transfer learning methods have shown great success in text classification tasks with limited data.
Our proposed framework can be seen as a specialized form of transfer learning, in which embeddings are imported from related, rather than general purpose, tasks or datasets, and in which several such embeddings are combined to form a single ensemble.

\subsection{Item categorization tasks}

Item categorization is a widely studied classification task which aims at classifying e-commerce products into the most suitable category (from a predefined set of categories) relying on the product's attributes, such as title, description, manufacturer, price, etc.

Most of the studies in this field have focused on the textual attributes (i.e., title and description) and treated the problem as a text classification task.
Earlier studies relied on representing text as a vector in a multidimensional space of identifiers (e.g., single index terms or N-grams) using various weighting schemes (e.g., TF-IDF).
For example, \cite{ding2002goldenbullet} compared several classification algorithms for product classification (e.g., KNN and Naïve Bayes).
They also compared the usefulness of using hierarchical classification methods vs. flat classification of all categories.
\cite{yu2012product} compared the effect of stemming and stop word removal pre-processing methods and different feature representations of words (e.g., Binary and TF-IDF) with a linear SVM classifier.
They also compared different multi-class classification strategies (e.g., one vs. all and one vs. one).
\cite{mathivanan2018improving} examined the effect of first clustering the item data using a K-Means algorithm, and then build a supervised KNN model separately for each cluster.
\cite{sun2014chimera} employed a combination of machine learning classifiers, rules created by analysts, and crowdsourcing in order to maintain an accurate item categorization system in production.
However, these methods share the following limitations:
1) they typically produce training matrices which are very high dimensional and consequently very sparse;
2) the semantic meaning of words and their relationship to other words is typically ignored; and
3) the context of words (e.g., where they appear in the sentence, after which word, etc.) is rarely taken into account.

With the growing popularity of deep learning, studies have started to show the applicability of deep learning also in the domain of item categorization 
\cite{das2016large} compared the accuracy of a linear model, a gradient boosting model and a CNN model in the item categorization task, and suggested a semi-automatic filtering method of noisy data using an unsupervised topic model.
\cite{ha2016large} used multiple RNN layers and fully connected layers for encoding product textual attributes in a deep neural network on a large scale item categorization task.
They also studied the effect of different parameters and different metadata on the classification accuracy.
\cite{cevahir2016large} used deep belief nets and deep autoencoders in order to classify item descriptions in Japanese into a very large set of categories.
\cite{li2018don} used a sequence to sequence architecture inspired by machine translation neural architectures with RNNs and Transformer layers to produce the whole product taxonomy with one model instead of using hierarchical classification or flat classification methods. 
\cite{krishnan2019large} used LSTMs and CNN layers for both text and non-text product attributes.
They investigated the effect of different neural architectures and different ways to combine structured product attributes in the deep neural network.
\cite{chen2019fine} used Character-level Convolutions with skip connections to predict the item category and determine whether a word in the item description is a category.
However, a major limitation of deep learning methods is the need for very large labeled datasets to train on.

Recent studies have tried to reduce the size of labeled dataset that is needed for training by using general purpose embeddings and general purpose pretrained models.
For example, \cite{kozareva2015everyone} compared several feature representations of text (TF-IDF with N-grams, Latent Dirichlet Allocation topic modeling, Mutual Information features and Average Word2Vec vectors) followed by a linear classifier.
This work was one of the firsts to use embeddings from neural networks (Word2Vec) in an item categorization task, and to demonstrate the benefits of using such an approach.
\cite{verma2018deep} and \cite{leecbb2020CBB} used CamemBERT, a pre-trained language model in french, in the e-commerce multi modal item categorization challenge as part of SIGIR 2020.
Specifically, CamemBERT was used as a text encoder, and was combined with a CNN that was used as an image encoder.
\cite{zaheraprobert} and \cite{Yang2020BertWD} used a finetuned version of BERT for item categorization on the ``semantic web challenge on mining the product data in websites'' (MWPD2020).

\section{Datasets}
\label{datasets}

\subsection{Invoice Dataset}
\label{Inv_dataset}

Our main dataset contains 3,461,897 invoices from PayPal's P2P transaction service, sampled and provided to us by PayPal’s team.
The invoices were sampled randomly from all invoices, having a seller from the US, and a creation date between September 15 - October 13, 2018.
Each transaction in our dataset contains the following fields: item name, item description, price, quantity of items, Buyer ID, seller ID and seller's industry.
Table \ref{tab:invExample} presents a few representative examples of such invoices.

\begin{table}[h!]
\begin{center}
\caption{Example of six PayPal's invoices and their attributes}
\label{tab:invExample}
\resizebox{\textwidth}{!}{
\begin{tabular}{|c|c|c|c|c|c|c|c|} 
\hline
\textbf{Invoice ID} & \textbf{Item name} & \textbf{Item description} & \textbf{Creation date}  & \textbf{Price} & \textbf{Seller's industry} & \textbf{Seller ID} & \textbf{Buyer ID} \\
\hline
1 &  \makecell{Orange and red \\ bralette large} &  & 2018-09-18  & 23.15 & fashion & 222 & 42 \\
\hline
2 & Voice Over Recording & \makecell{Recorded the part of \\ Michael the CEO for the entirety \\ of the project.} & 2018-10-12  & 1000 & media & 15 & 765 \\
\hline
3 & OS Leggings &  & 2018-09-07 & 38 & fashion & 12 & 899 \\
\hline
4 &  \makecell{Transport to north ireland \\ for ebay pellet hose} &  & 2018-10-01  & 32.62 & auto-parts & 65 & 78 \\
\hline
5 & Product &  & 2018-09-18 & 81 & photography & 132 & 65 \\
\hline
6 & Bunnies & 1/2 yard & 2017-11-12 & 27.75 & arts-n-craft & 2 & 555 \\
\hline
\end{tabular}
}
\end{center}
\end{table}

\begin{figure}[H]
\centering
\includegraphics[width=0.9\linewidth]{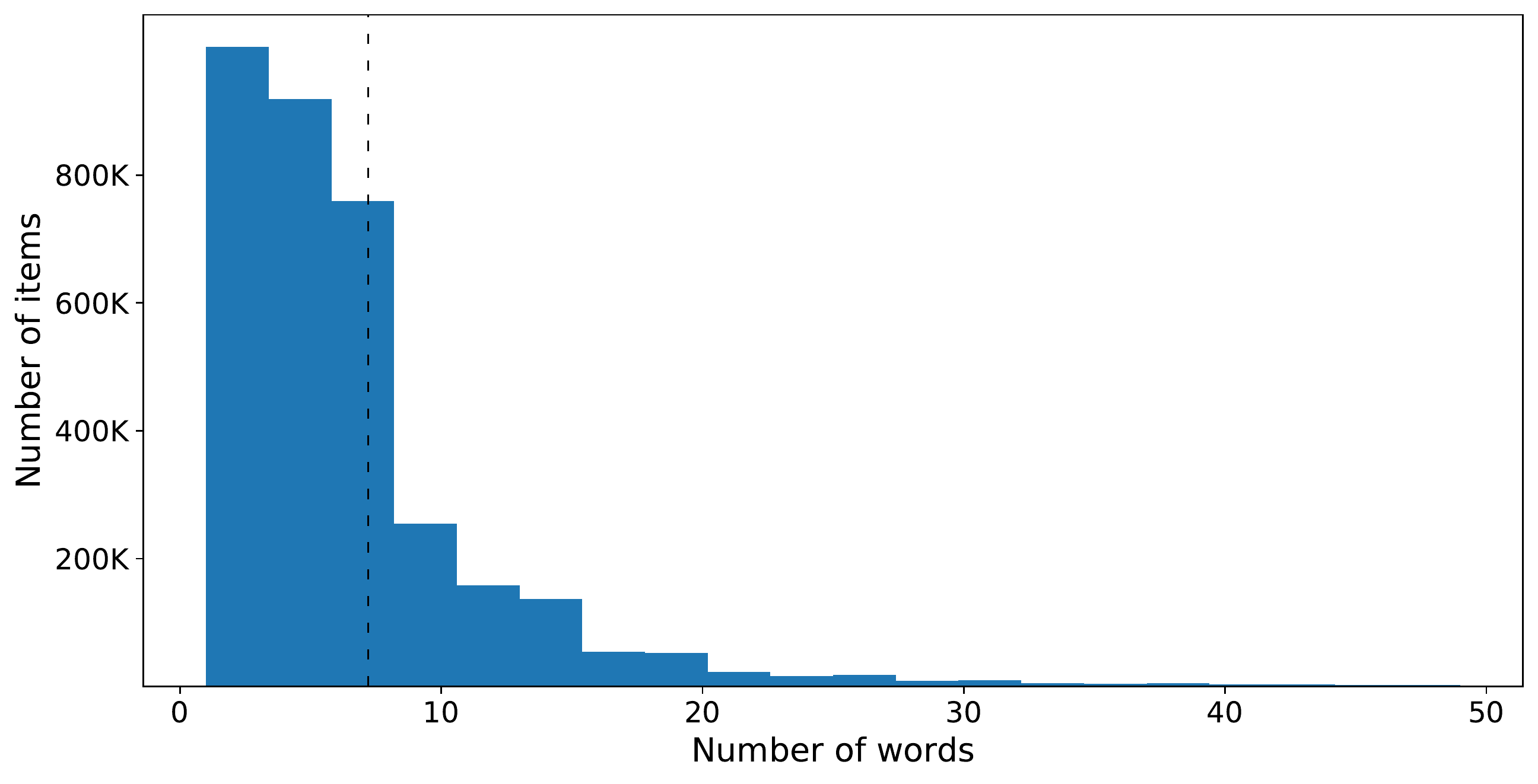}
\caption{Histogram of the number of words per item description in the invoice dataset. The vertical dashed line represents the mean number of words per item description.}
\label{InvLen}
\end{figure}

The item name and description fields are free text fields that are filled by the seller without any constraint nor validation.
Thus they are typically very short and noisy\footnote{We use the term noisy to describe user generated text that typically contain grammatical errors, nonstandard spellings, abbreviations, etc., as previously done with tweets on Twitter, \citep{baldwin2015shared}} (see for example, invoices 1 and 3 in Table \ref{tab:invExample}), and sometimes not informative at all with regard to the sold item nor its category (see for example invoices 5 and 6 in Table \ref{tab:invExample}).
It should also be noted that item description is an optional field and therefore is not necessarily filled by the seller.
In fact, it contains a value only in 24\% of the invoices in our dataset. 
Figure \ref{InvLen} presents a histogram of the number of words per item (after concatenating the item name and description fields).

\begin{figure}[H]
\centering
\includegraphics[width=0.9\linewidth]{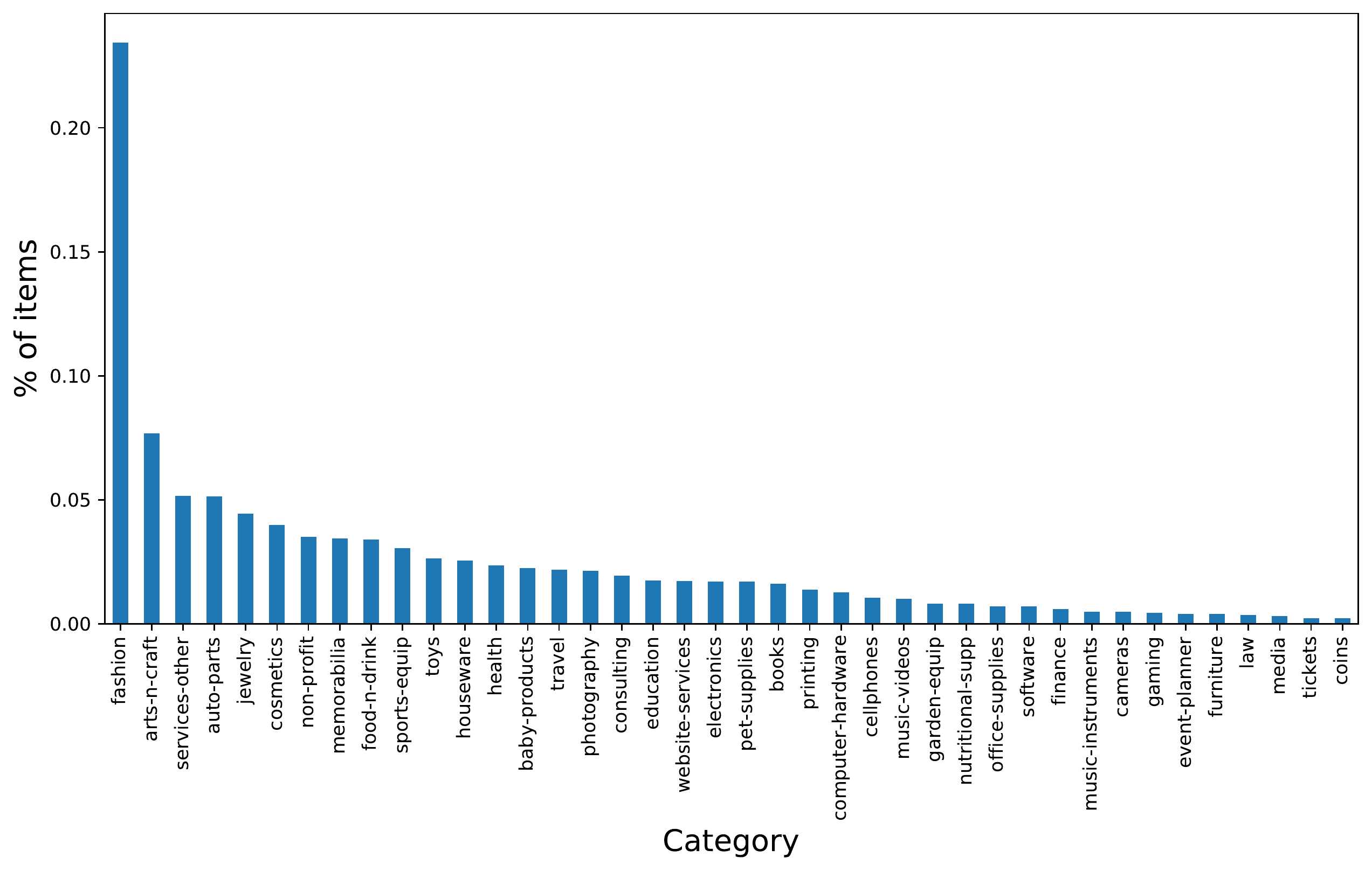}
\caption{The distribution of items by seller's industry in the invoice dataset.}
\label{InvCats}
\end{figure}

The seller's industry field represents a classification of the seller itself (i.e., not a specific invoice/item) into a single category out of 40 predefined categories (such as fashion, jewelry, accounting, etc.).
(The goal of this paper is to classify the item sold in an invoice into one of these 40 categories.)
This field is filled in the vast majority of cases automatically using PayPal's proprietary algorithm.
This field is highly imbalanced, where the most popular category (fashion) contains 24\% of the invoices, and the 10 most popular categories contain 78\% of the invoices.
A histogram of the various categories is presented in Fig. \ref{InvCats}

It should be noted that before providing us the invoices, PayPal performed an initial filtering procedure which originally involved roughly 8 million invoices, and filtered-out invoices with the following properties:
\begin{itemize}
\item Item description is shorter than 3 symbols.
\item Item description contains a single token which is one of the following: ``invoice'', ``order'' and ``unknown''.
\item Invoice items with duplicate values in the text attributes were kept once.
\end{itemize}
The 3,461,897 invoices analyzed in this study are those remaining after the initial filtering procedure.

\subsection{eBay Dataset}
\label{ebay_dataset}

This dataset contains 2,997,571 items that were extracted using eBay's public API.
The items were sampled randomly from all items on the eBay US website (eBay.com), having a creation date between September 1 - October 26, 2018.
We only used items that were sold at least once.
Each extracted item contains the following fields: item description, country, price, creation date, and category.
Table \ref{tab:ebayExample} presents a few representative examples of such items.

\begin{table}[h!]
\begin{center}
\caption{Example of five eBay items and their attributes}
\label{tab:ebayExample}
\resizebox{\textwidth}{!}{
\begin{tabular}{|c|c|c|c|c|c|} 
\hline
\textbf{Record ID} & \textbf{Item description} & \textbf{Country} & \textbf{Price} & \textbf{Creation date} & \textbf{Category} \\
\hline
1 & \makecell{"VINTAGE STERLING SILVER 20"" long  FINE ROPE \\  LINK NECKLACE CHAIN - 3g!"} & GB & 24.38 & 2018/10/15 & jewelry \\
\hline
2 & \makecell{Off White Red iPhone X SE 5 6 7 8 S Plus \\ Off White iPhone Case  [For Apple iPhone 8 Plus]} & ID & 22.98 & 2018/09/20 & cellphones \\
\hline
3 & \makecell{Superhero Smash Hands Gloves Ironman spiderman \\  Hulk The Avengers 1 Pair new [hulk]} & C2 & 11.16 & 2018/09/04 & toy \\
\hline
4 & \makecell{SWIFTLET Tempered Glass Screen Protector for \\ iPad 2 3 4 5 6 Air Mini Pro 9.7 [iPad Air 1/2]} & US & 13.24 & 2018/09/14 & arts-n-craft \\
\hline
5 & \makecell{Real Premium Tempered Glass Film \\ Screen Protector For Apple iPad 1/ 2 / 3 / 4} & SG & 25.07 & 2018/10/14 & furniture \\
\hline
\end{tabular}
}
\end{center}
\end{table}

\begin{figure}[H]
\centering
\includegraphics[width=0.9\linewidth]{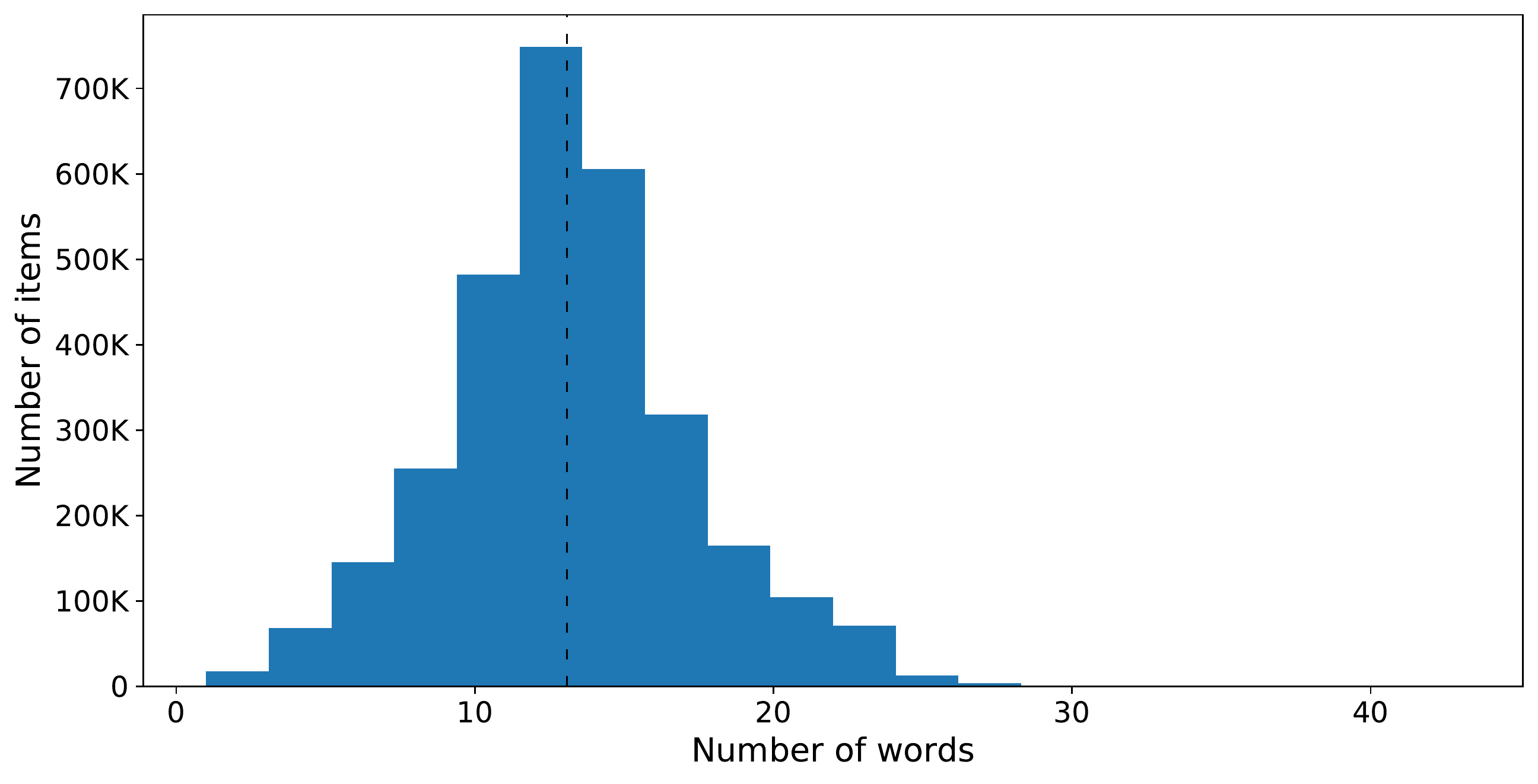}
\caption{Histogram of the number of words per item description in the eBay dataset. The vertical dashed line represents the mean number of words per item description.}
\label{ebayLen}
\end{figure}

\newpage
The item description field in this dataset is significantly longer and less noisy than that of the invoices dataset (see for example, record 1 in Table \ref{tab:ebayExample}).
This is in part since it is used to retrieve items in eBay's item search.
Moreover, eBay encourages sellers to augment their item descriptions with relevant keywords, which makes the description of two related items quite similar. 
For example, records 4 and 5 in Table \ref{tab:ebayExample} both share the keywords Tempered Glass, Screen Protector, and iPad.
Figure \ref{ebayLen} presents a histogram of the number of words per item description.

\begin{figure}[h!]
\centering
\includegraphics[width=0.9\linewidth]{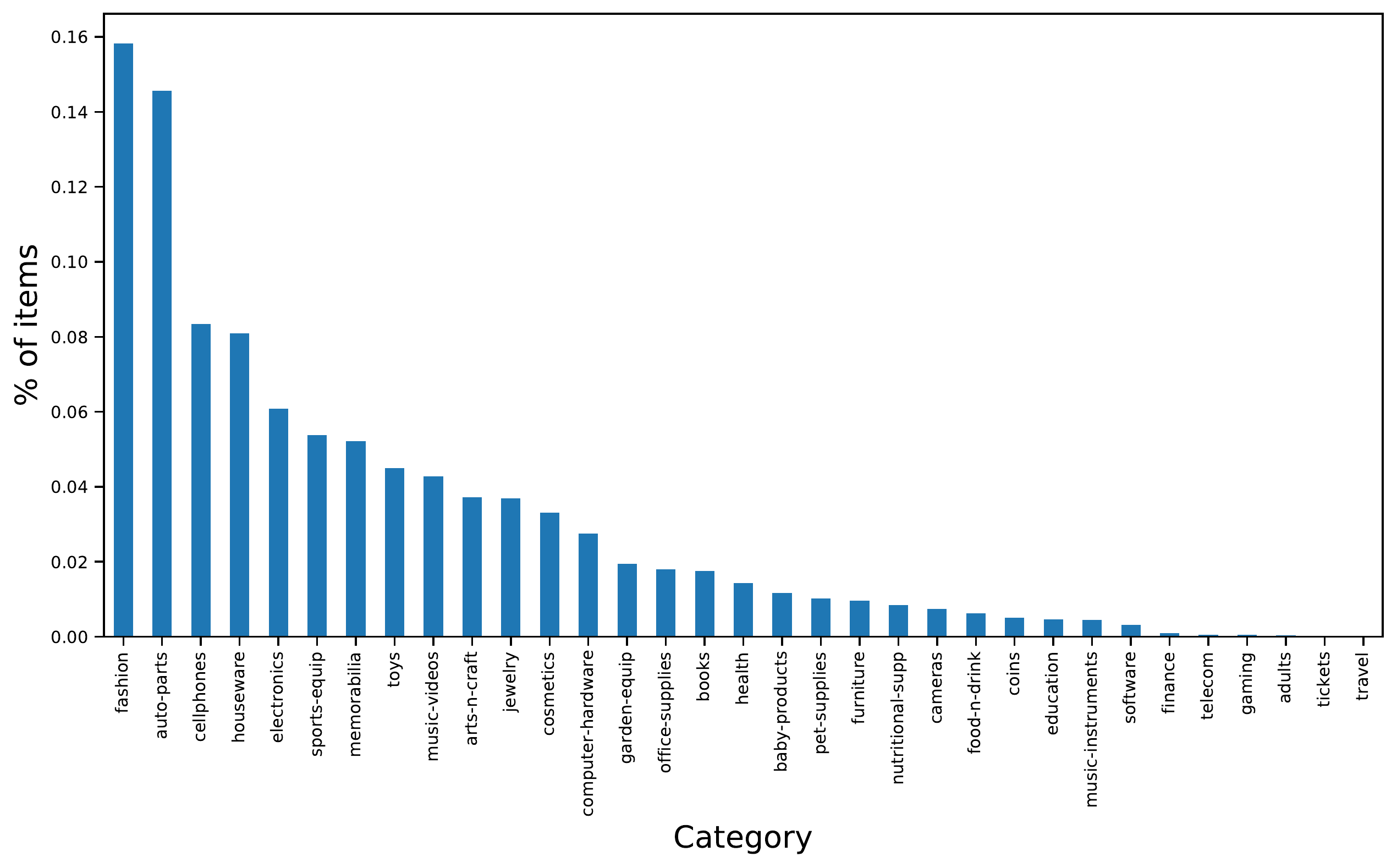}
\caption{The distribution of items by category in the eBay dataset.}
\label{ebayCats}
\end{figure}

The category field here classifies an item into one of eBay's predefined set of categories.
It is important to note that the set of eBay categories is different from the set of PayPal categories.
First, the set of eBay categories contains thousands of categories while the set of PayPal categories contains only 40 categories.
Second, eBay categories are limited to goods (i.e., physical items), whereas PayPal categories include also services.
To overcome this difference, eBay categories were mapped to their corresponding PayPal categories (33 out of the 40 PayPal categories that represent goods) using a set of rules determined manually by PayPal's domain experts.
A histogram of the resulting set of categories is presented in Fig. \ref{ebayCats}

\section{The proposed method}
\label{method}

As stated in section \ref{intro}, deep neural networks are currently the state-of-the-art approach for text classification in general, and for item categorization in particular.

Deep neural networks generally need a huge amount of labeled data in order to not overfit the training set.
For example, \cite{krishnan2019large} used 23 million manually labeled items for training their item-categorization deep neural network.
Manually labeling such a big dataset may be extremely expensive and time consuming, even when using crowdsourcing.

Another approach to leverage neural networks without needing huge labeled datasets, is by using previously learned general purpose embeddings or pretrained models, such as Word2Vec \citep{mikolov2013distributed} or BERT \citep{devlin2018bert}.
These embeddings were generated once by a neural network that was trained on a huge general purpose text corpus (e.g., Wikipedia, books, and news articles), using a self supervised target (e.g., the next word in a sentence).
After these embeddings were learned, they can be used as input features for much smaller labeled datasets.

While general purpose embeddings have demonstrated great value for various text classification tasks, we suspect that they may not be optimal in cases where the target corpus's text distribution differs considerably from that of the general-purpose corpus they were trained with.
Recall that in our setting, item descriptions contain very short, noisy, and domain specific text, thereby presenting a very different text distribution from that of general purpose corpora.

To address this challenge, we propose the Ensembled Transferred Embedding (ETE) learning framework.
The ETE framework first manually labels a small sample of instances from the target dataset and generates embeddings from related large-scale datasets.
Then, each set of generated embeddings and the labelled instances of the target dataset are used to train a model that is tailored to the target task.
Finally the set of trained models are combined into a single ensemble to provide enhanced performance.
This motivation behind this approach is twofold:
(1) using related (or ``transferred'') embeddings as opposed to general purpose embeddings makes the resulting model more tailored to the target task, and
(2) generating the embeddings from related datasets bypasses the need to label a large portion of the target dataset. 

In the rest of this section, we first provide more details on the proposed ETE learning framework, and then describe the required details for applying this framework to our unique item categorization task.

\subsection{Ensembled transferred embeddings}

\begin{figure}[H]
\centering
\includegraphics[width=0.9\linewidth]{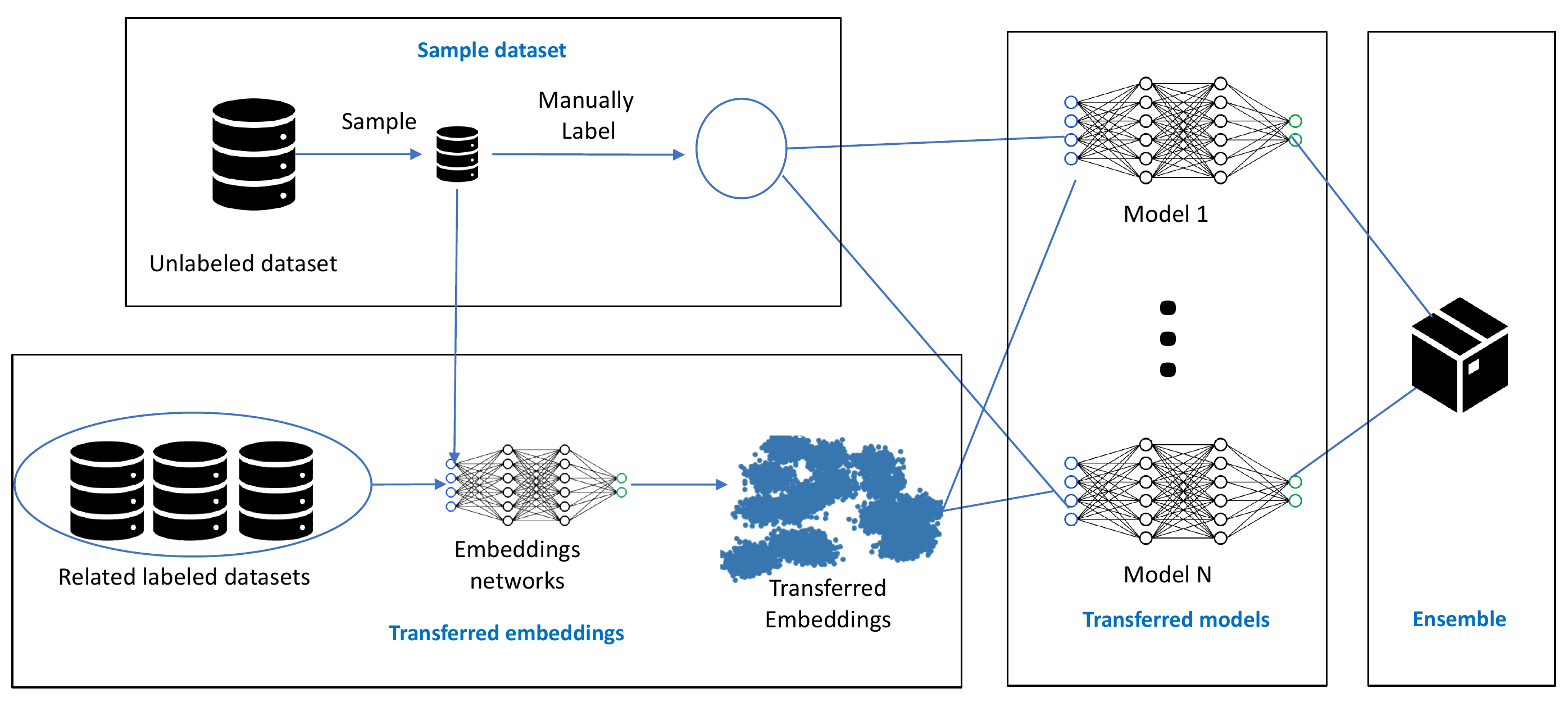}
\caption{High-level architecture of the ETE framework.}
\label{GenSchema}
\end{figure}

Our Ensembled Transferred Embeddings framework is composed of four main steps (see Figure \ref{GenSchema}):

\begin{itemize}
\item Step 1: Sample Dataset. In this step, we generate a relatively small-scale manually labeled dataset.
This may be achieved by randomly sampling a relatively small number of instances from the large-scale unlabelled target dataset, and manually label these instances.
Labeling can be done by using domain experts, crowdsourcing, heuristics, or a combination of the above.

\item Step 2: Transferred Embeddings. In this step, we extract embeddings from related large-scale labeled datasets.
To achieve this goal, we first need to identify related datasets.
Such datasets are large-scale labeled datasets from a domain or task, similar in nature to that of the target dataset.
Such datasets may include, for example, a publicly available dataset with a similar task from a different domain, a dataset that was gathered for a different task on the same domain, or even a self supervised task on the target dataset.
After the related datasets were obtained, a deep neural network is trained on them to obtain an ``embedding network''.
Finally, the sample dataset is provided as input to the embedding network to generate ``transferred embeddings'', in a manner similar to \citep{sharif2014cnn} and \citep{kiros2015skip}.

\item Step 3: Transferred Models. In this step, we train models using the extracted transferred embeddings and the labels of the sample dataset.
More specifically, for each set of transferred embeddings and their corresponding labels (from the sample dataset), we apply a supervised machine learning algorithm to obtain a ``transferred model''.

\item Step 4: Ensemble. In this step, a meta-model for combining the outputs of the various transferred models is built.
This can be obtained in several ways, for example by using domain knowledge, applying a voting rule such as plurality (see the work by \cite{werbin2019beyond} for more details), or by training a meta-model to learn the right combination method.

\end{itemize}

Given a new data instance we first use the trained embedding networks to extract transferred embeddings.
We then apply the transferred models on the extracted transferred embeddings.
Finally, we combine the outputs of the various transferred models using the ensemble to decide on the predicted category.

\subsection{Applying ETE to our item categorization setting}




In this section, we describe the required details for applying the ETE framework to our unique item categorization setting.
Specifically, we elaborate on how the sample dataset was extracted and labeled; how the transferred embeddings were obtained; how the transferred models were trained, and how the ensemble was built.

\subsubsection{Sample Dataset - Labeling using MTurk}
\label{subsec:mturk}

Our sample dataset was obtained by a uniform sampling of 1970 instances\footnote{The specific number of 1970 instances was chosen to fit our budget constraint of 200 USD.} from the invoices dataset and manual labeling of these instances using Amazon Mechanical Turk (MTurk) as we proceed to explain.

Amazon Mechanical Turk is a marketplace for crowd-sourcing of Human Intelligence Tasks (HITs) that is often used to generate labels for supervised machine learning tasks.

Each of our 197 MTurk tasks was composed of labeling a set of 10 instances (1970 instances in total) and was sent to 5 different workers (Turkers).
Each task started by providing a general description of the study, followed by a detailed explanation of how the questions should be approached, and a detailed description for each of the 40 categories (including examples of items that fit that category).

\begin{figure}[H]
\centering
\includegraphics[width=1.0\linewidth]{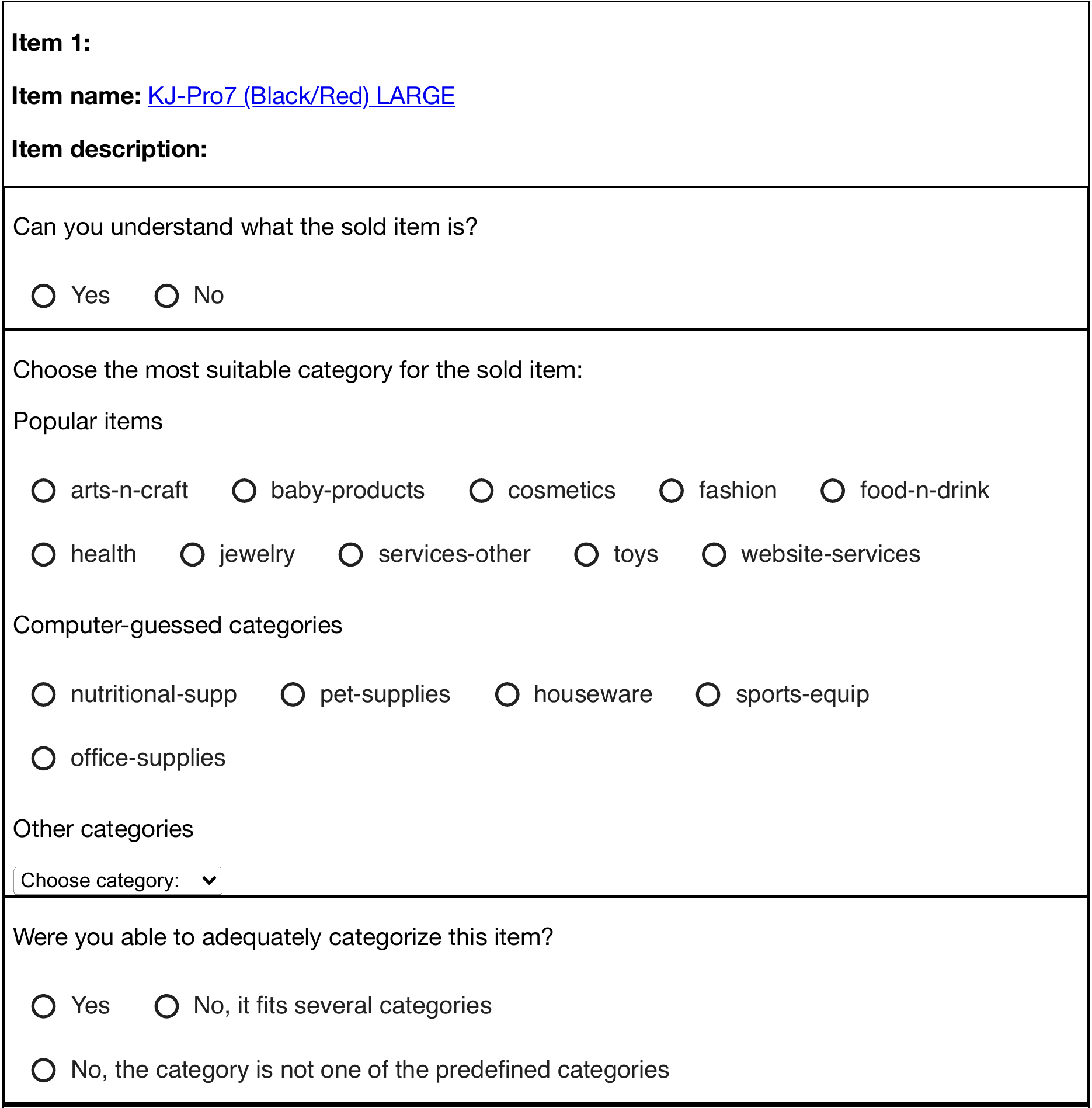}
\caption{An example of a single instance to be categorized as appeared in the Mechanical Turk task}
\label{mturk}
\end{figure}

Then, the Turker was asked to answer the task.
For each of the 10 instances appearing in the task, we provided the Turkers with the item name and description and asked them to answer the following three questions:
\begin{enumerate}
\item Can you understand what the sold item is? [yes/no]
\item Choose the most suitable category for the sold item [from a list of provided categories]
\item Were you able to adequately categorize this item? [yes/no with explanation]
\end{enumerate}
An example of such a task is presented in Figure \ref{mturk}.

It is important to note again that the set of PayPal categories includes 40 categories, and requesting the Turker to go through a list of 40 categories to choose the one that best fits the item description is a tedious task.
Indeed, in preliminary experiments that we conducted, we found that providing all 40 categories as options led to a high level of disagreement between Turkers and to a high level of bias towards the categories shown at the beginning of the list.

To overcome this issue, we decided to split the list of categories into 3 parts:
1) most popular - the 10 most popular categories, according to the seller's industry attribute, presented in alphabetical order;
2) computer guessed - the 5 most probable categories (that are not among the 10 most popular categories) according to some baseline model (Logistic regression on TF-IDF representation with industry as the label); and
3) other categories - the rest of the 40 categories.
Clearly, while addressing the bias mentioned above, this design could lead to a different type of bias, by which items from the most popular or computer guessed categories would have higher likelihood to be chosen.
Nevertheless, we believe that such a bias is minor since turkers still chose categories from the ``other categories'' list in a non-negligible number of cases, and the overall level of agreement between turkers increased considerably.

In order to further help Turkers categorize the items in a fast and accurate way, we provided them a hyperlink to a Google search of the item name, and for each category in the list of categories, we provided a tooltip that included the category's description (the short description of the category that also appeared in the instructions page).

After completing the data collection stage, we reviewed the data to remove ``lazy Turkers'' (i.e., Turkers that have not spent enough time and effort to answer the tasks).
We did so by first calculating the level of agreement for each instance (i.e. the maximum number of Turkers that have assigned the same category to that instance).
We then identified instances with a level of agreement between 3 to 5.
Finally, we removed Turkers (and their answers) that agreed with the most-agreed category in less than 20\% of these instances (and all of their answers).
Since we used Turkers with a master title (required additional payment to MTurk), we found only two such "lazy Turkers''.
After their removal, we sent all of their tasks to MTurk to obtain new and valid answers.

Finally, we used the resulting data to label the instances.
If 3-5 Turkers answered "no" to the first question of a given instance, that instance was labeled as uninformative (15\% of the instances).
If 3-5 Turkers agreed on the category, this category was assigned as the label for that instance (75\% of the instances).
Finally, the rest of the instances were reviewed manually by a domain expert in order to choose the best fitting category (10\% of the instances).

\subsubsection{Transferred Embeddings - using Seller's Industry Attribute and eBay Dataset}

\begin{figure}[H]
\centering
\includegraphics[width=0.9\linewidth]{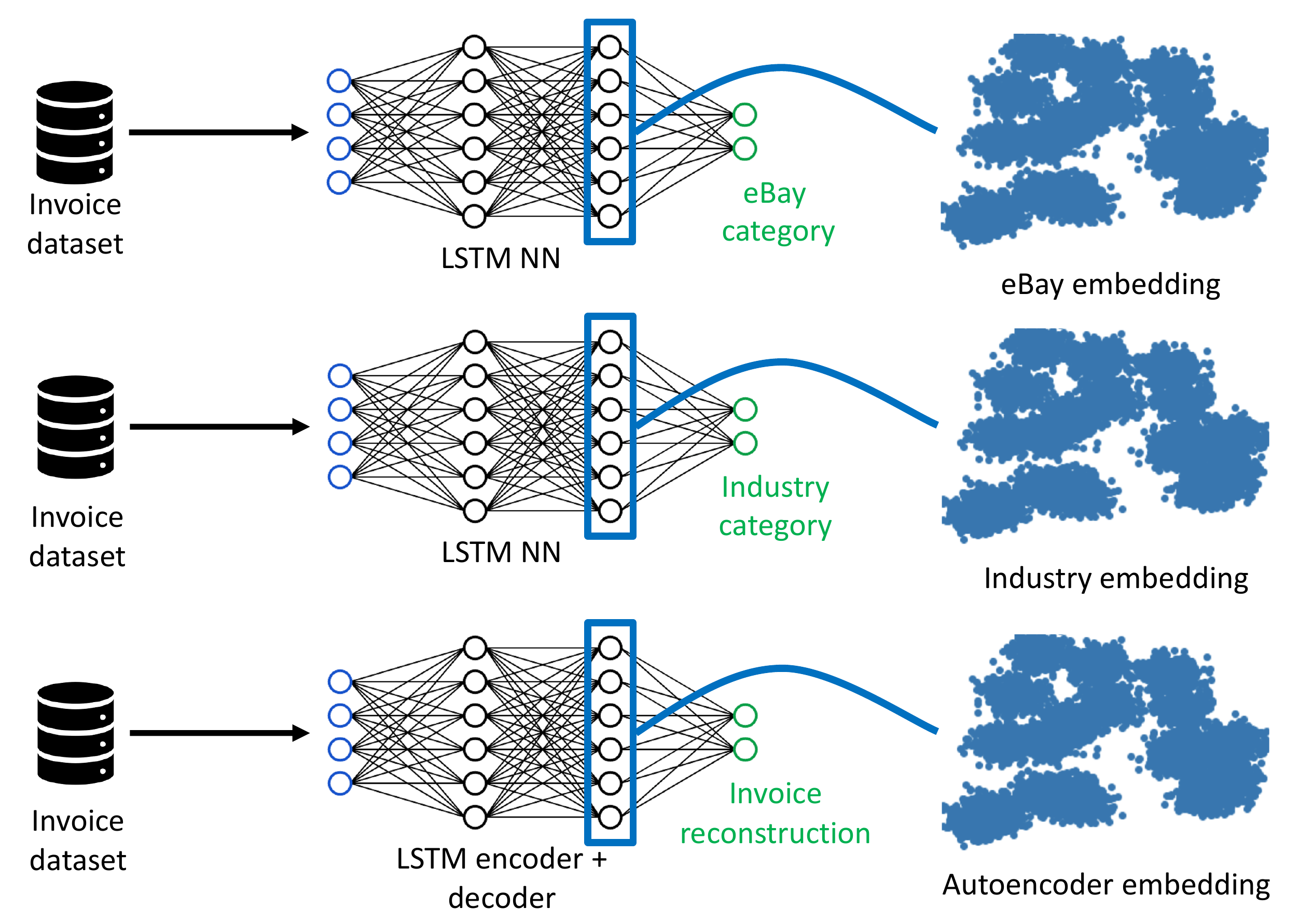}
\caption{The process for generating transferred embeddings in our setting.}
\label{EmbSchema}
\end{figure}

\smallskip
\noindent
We use three types of transferred embeddings (as illustrated in Figure \ref{EmbSchema}):

\begin{itemize}

\item \textbf{Industry embedding}.
\label{subsec:ind_emb}

Recall that we have access to a large number of invoices, but none of these invoices are labeled (i.e., the right item category for each invoice is unknown).
Fortunately, recall that each invoice is associated with the seller's industry category.
While the seller's industry category is clearly related to the item category (fashion website will mostly sell fashion items), the two categories are not identical for two main reasons:
1) sellers usually sell more than one category of items (e.g., a fashion website can also sell jewelry or baby clothes), and
2) the seller's industry category itself may be wrong (since it was generated by an automated algorithm).
To further support this claim, an analysis of the MTurk dataset shows that the seller's industry category matches the item category in 48\% of the cases only.
To conclude, since the seller's industry category can be seen as a noisy label for the item category, we train an LSTM neural network to predict the seller's industry category from the item description and use the last layer before SoftMax as the industry embedding (see more details below).
It is important to note that due to the differences between goods and services categories we decided to split the industry embedding into two types of embeddings: a goods embedding and a services embedding.
Each of the two embeddings was generated by training the LSTM neural network only on part of the invoices data (either those that have a seller's industry category associated with goods or those that have a seller's industry category associated with services).

\item \textbf{eBay embedding}.
\label{subsec:ebay_emb}

In contrast to the invoice dataset, the eBay dataset has reliable item category labels, but it differs from the invoice dataset in two ways.
First, the eBay dataset contains only a subset of the potential invoice categories since eBay sells goods while our invoices, which are the result of P2P transactions, contain also services.
Second, the item description in eBay is longer, more standardized, and less noisy than the item description in the invoice dataset, since it is aimed to be found in the eBay search.
To conclude, since in both cases we aim at predicting the item category from the item description, we train an LSTM neural network to predict the eBay's item category from the eBay's item description and use the last layer before SoftMax as the eBay embedding (see more details below).

\item \textbf{Autoencoder embedding}.
\label{subsec:auto_emb}
The last type of embedding is an embedding trained using a self-supervised Autoencoder.
An Autoencoder is a model that is trained to reconstruct its own input.
We use a similar LSTM architecture to the two neural networks mentioned above for the encoder (without softmax) and a mirror architecture with a fully connected layer in the end for the decoder.

\end{itemize}

\noindent
The three LSTM deep neural networks mentioned above share the same architecture and hyper-parameters.
We chose to use LSTM deep neural networks due to their successful implementation in previous works on item categorization tasks \citep{krishnan2019large, li2018don}.
The chosen architecture and hyperparameters are detailed in Section \ref{results}.
The rationale for using the exact same architecture and hyperparameters in the three networks was merely a matter of simplicity\footnote{Our goal here was to demonstrate the advantages of the ETE framework on a large-scale real-world problem, rather than pursuing the best possible accuracy.}.

\subsubsection{Transferred Models - using a fully connected neural network}

After generating the transferred embeddings described above, we turn to train ``transferred models'', each trained using a different transferred embedding and the Mturk dataset.
More specifically, given a train set of instances extracted from the MTurk dataset, for each transferred embedding (separately), we generate a new training set.
The features of the new training set are generated by applying the transferred embedding on the features of the original MTurk instance, and the target value is simply copied as is from the original MTurk instance.
The resulting set of instances is then used to train a trasferred model.
For this purpose we use a relatively small fully connected neural network with 1 layer of 100 hidden units and a softmax layer.

\subsubsection{Ensemble - Stacking models}

To combine the results of the four transferred models, we use a common ensemble technique called model stacking \citep{wolpert1992stacked}.
More specifically, given a train set of instances extracted from the MTurk dataset, we apply each of the transferred models to obtain a prediction (a features vector containing a single probability value for each of the 40 possible categories).
Next, a new instance is produced by unifying the predictions into a single features vector and copying the target value as is from the original MTurk instance.
Finally, the resulting instances are used to train a Logistic Regression meta model.
This process allows our stacking meta model to learn the strengths and weaknesses of each transferred model and consequently lead to improved predictions.



\section{Evaluation}
\label{results}

In this section we report the extensive evaluation that we conducted.
We begin by describing the experimental setting in subsection \ref{subsec:exp_setting}, followed by the results in subsection \ref{subsec:results}.
The source code used for the proposed framework and its evaluation can be found in \citep{sourcecode2021}.

\subsection{Experimental setting}
\label{subsec:exp_setting}

The ETE method was evaluated by comparing its classification performance to that of seven other benchmark methods (see Section \ref{subsec:compare_methods}) over the labelled MTurk dataset (see Section \ref{subsec:dataset_pre_precossing}).
We applied a stratified 10-fold cross validation evaluation scheme, where in each iteration of the cross validation process both the transferred models and the ensemble were trained on the nine training folds and tested on the test fold.
We used accuracy and weighted F1 score\footnote{The harmonic mean of precision and recall of each class weighted by the class proportion in the data} as measures for classification quality.
To avoid data leakage from the training set to the test set, we removed all the samples that were used in the Mechanical Turk dataset from the invoice dataset.

\subsubsection{Compared methods}
\label{subsec:compare_methods}

We compare our method to four common item categorization methods and three methods based on transferred embeddings.

\begin{itemize}
\item \textbf{Majority:} Our first method is taking the most frequent category in our dataset (fashion) as the predicted class.

\item \textbf{TF-IDF:} Our first baseline model is a regularized logistic regression model trained on TF-IDF with unigrams and bi-grams, similar to \citep{kozareva2015everyone}. The TF-IDF was built using the 3500 most popular terms in the invoice dataset (see Figure \ref{tfidf_terms} in the appendix for more details on why the number 3500 was chosen).
\item \textbf{Average GloVe:} A fully connected neural network trained on average pre-trained GloVe word embedding as input vectors, similar to \citep{kozareva2015everyone}.
\item \textbf{BERT:} A fully connected neural network trained on embeddings extracted from BERT base uncased \citep{devlin2018bert}.
\item \textbf{Autoencoder:} A fully connected neural network trained on transferred embedding extracted from a self supervised auto-encoder \citep{erhan2010does} trained on the invoice dataset as described in Section \ref{subsec:auto_emb}.
\item \textbf{eBay embedding:} A fully connected neural network trained on the eBay transferred embedding as described in Section \ref{subsec:ebay_emb}.
\item \textbf{Industry embedding:} A fully connected neural network trained on the Industry transferred embedding as described in Section \ref{subsec:ind_emb}.
\end{itemize}

\subsubsection{Dataset and pre-processing}
\label{subsec:dataset_pre_precossing}

All of our experiments were conducted using the MTurk dataset.
The two text attributes, item name, and item description were then concatenated to a single text attribute.
The text attribute was then pre-processed by applying very basic text operations, such as tokenization, lower-casing, and removing non alpha-numeric symbols.

\subsubsection{Hyper-parameter tuning}
Recall that all three LSTM deep neural networks used for extracting embeddings shared the same architecture.
The chosen architecture and hyper-parameters were tuned using a grid search on the following hyper parameters: Number of LSTM layers (1-3), size of hidden dimension for the LSTM layers (100,200,400), size of hidden dimension for the fully connected layer (10,30,50,100) and dropout rates (0.1,0.2,0.3,0.5).
The chosen architecture and hyper-parameters are shown in Table \ref{table:architecture}.
We used Adam optimizer in all runs with a learning rate of 0.001.

\begin{table}[h!]
\begin{center}
\caption{Chosen architecture and hyper-parameters for the LSTM deep neural networks}
\label{table:architecture}
\resizebox{\textwidth}{!}{
\begin{tabular}{|c|c|c|c|} 
\hline
\textbf{Layer number} & \textbf{Layer} & \textbf{Size} & \textbf{Parameters} \\
\hline
1 & Embedding & 300 & Frozen pre-trained GloVe, max length=15 \\ 
\hline
2 & Spatial Dropout & & Dropout rate = 0.3 \\
\hline
3 & LSTM & 200 &  Return sequence = True\\ 
\hline
4 & LSTM & 100 & \\ 
\hline
5 & Dropout & & Dropout rate = 0.2 \\
\hline
6 & Fully Connected & 30 & Activation = relu \\ 
\hline
7 & Fully Connected & Number of classes & Activation = Softmax \\
\hline
\end{tabular}
}
\end{center}
\end{table}

\subsection{Results}
\label{subsec:results}

\begin{figure}[H]
\centering
\includegraphics[width=0.9\linewidth]{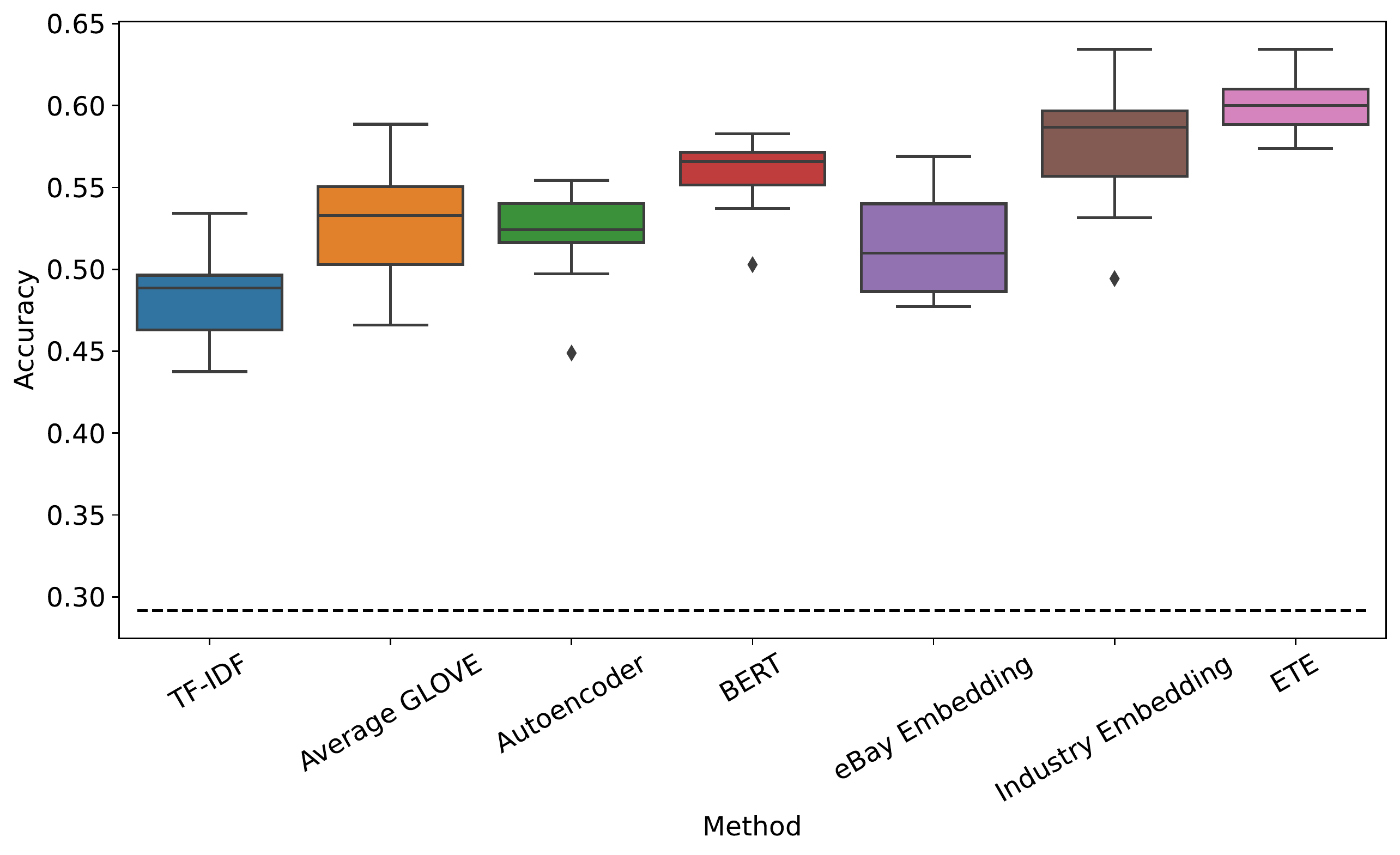}
\includegraphics[width=0.9\linewidth]{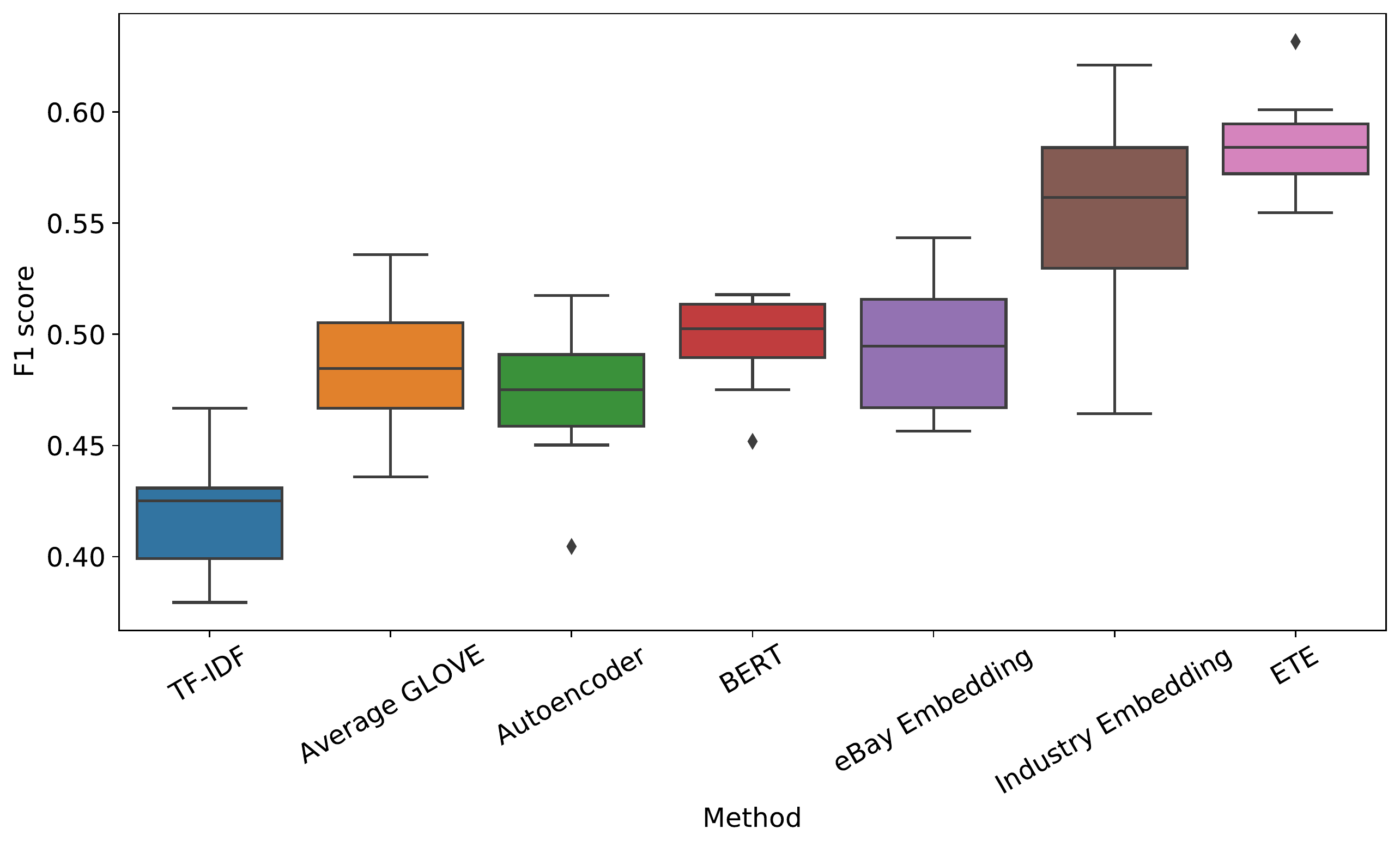}
\caption{A comparison of the different item categorization methods in terms of: Accuracy (top) and Weighted F1 score (bottom). The horizontal dashed line in the top subfigure represents the accuracy of the Majority method.}
\label{acc_bar}
\end{figure}

First, we computed the average classification accuracy and Weighted F1 score (over the 10 folds) for each one of the eight compared methods (see Figure \ref{acc_bar}).
As expected, all six embedding-based methods performed better than the basic TF-IDF methods.  
Moreover, we see that the Industry embedding method (which is based on transferred embedding) was able to outperform BERT which is the state-of-the-art general purpose embedding method.
We believe that this happens due to the unique characteristics of the text used in our setting --- item descriptions are short, noisy and are domain specific.
These characteristics make the text distribution in our setting different than that of the text used to train BERT, making BERT less suitable for our setting.
Finally, we see that ETE, which ensembles several transferred embeddings, performs the best in terms of classification accuracy and Weighted F1 score and has a low variance across folds compared to other methods.
While Figure \ref{acc_bar} presents the results for each of the Autoencoder, Industry Embedding and eBay Embedding methods alone, and where all three of them are combined together into a single ensemble, Figure \ref{partial_acc} in the appendix presents also the results for ensembles based on all pair combinations.

To further support our findings above, we performed statistical significance tests.
First, we performed an ANOVA analysis in order to reject the null hypothesis that all compared methods performed the same.
Second, we performed the Tukey post-hoc test to perform a pairwise comparison between the various methods.
With a confidence level of 95\%, we rejected the null-hypothesis that all methods performed the same. 
Further applying the Tukey post hoc test supported our finding that ETE performed significantly better than all other methods.

\begin{figure}[H]
\centering
\includegraphics[width=1.0\linewidth]{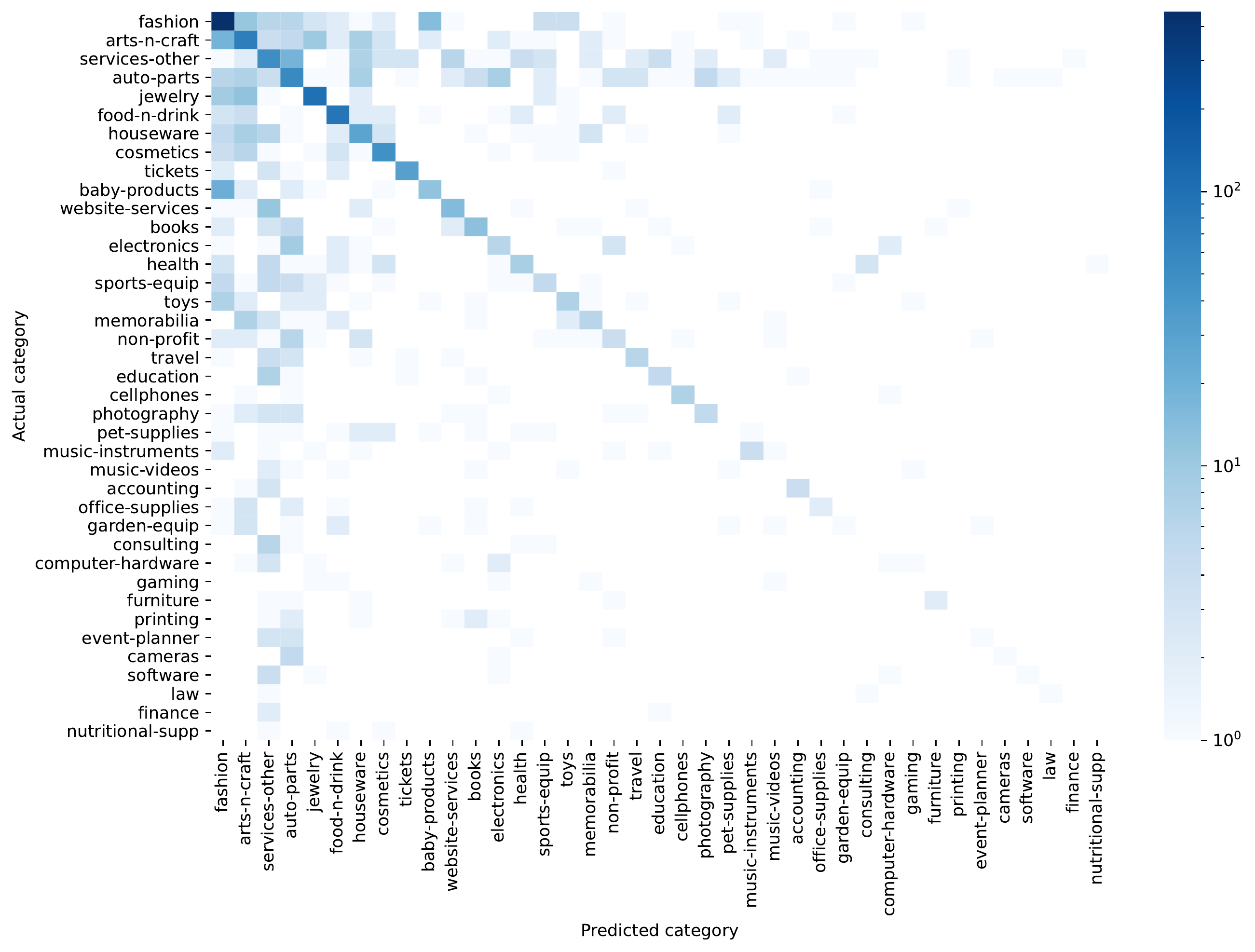}
\caption{The classification confusion matrix.}
\label{confusion}
\end{figure}

To better understand the cases in which our model had mistaken, we computed the classification confusion matrix (see Figure \ref{confusion}).
Note that categories on the two axes are sorted accordint to their overall popularity.
As expected, most of the model's classifications were successful, resulting in a relatively dark diagonal line.
Some worth noting confusions are between the fashion category and the baby-products category, and between the jewelry category and arts-n-craft category.
Although our evaluation considered such confusions exactly the same as a confusion between the electronics category and the food-n-drink category, it is clear that the former confusions are considerably more tolerable.

Figures \ref{len_comp} and \ref{tab:recall_improvments} in the appendix present two addition analyses of errors.
The first analysis examines the accuracy of the ETE model on different lengths (words count) of item descriptions.
The second analysis examines the improvement of the ETE model on the 15 most common categories compared to the industry model which seems to be our strongest baseline.

\begin{figure}[H]
\centering
\includegraphics[width=0.9\linewidth]{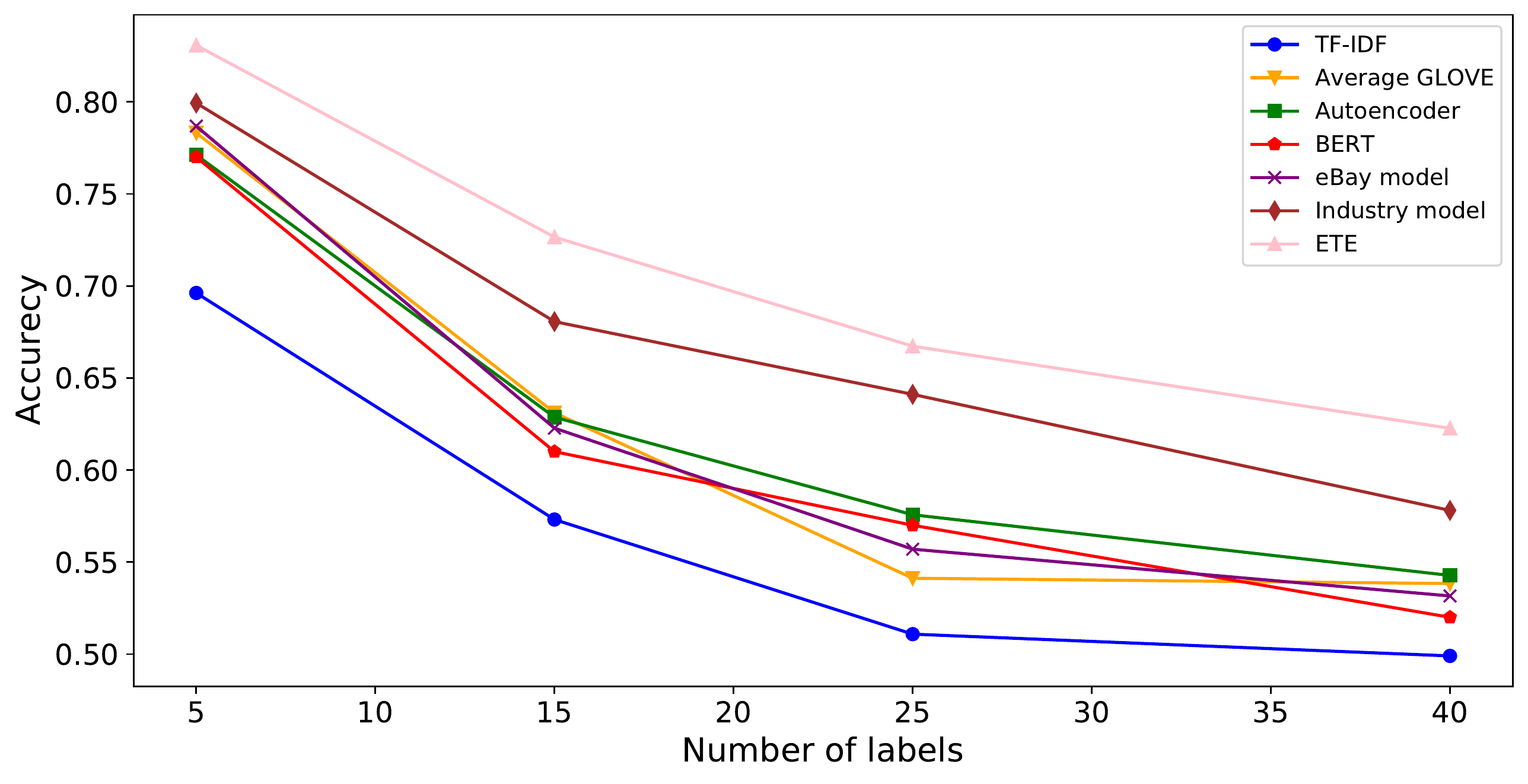}
\caption{Accuracy as a function of the number of categories considered.}
\label{Num_categories}
\end{figure}

Since the number of PayPal categories is relatively large, the likelihood of our model (and in fact any model) to confuse between them is high.
This is especially true for rare categories for which the model had a small number of instances to train with.
In Figure \ref{Num_categories}, we demonstrate the impact of restricting the number of categories to the top X most common categories (for X ranging between 5 and 40) and ignoring all instances that belong to the other categories, on the accuracy of the various compared methods.
As expected, we see that all methods performed considerably better when restricted to a lower number of categories. 
Moreover, we see that the proposed ETE method, outperforms the other method in all considered cases.
Figure \ref{num_categories_with_other} in the appendix presents a similar analysis in which the rest of the categories are not ignored but rather joined into a single ``other'' category.

\begin{figure*}[t!]
\centering
\includegraphics[width=0.8\textwidth]{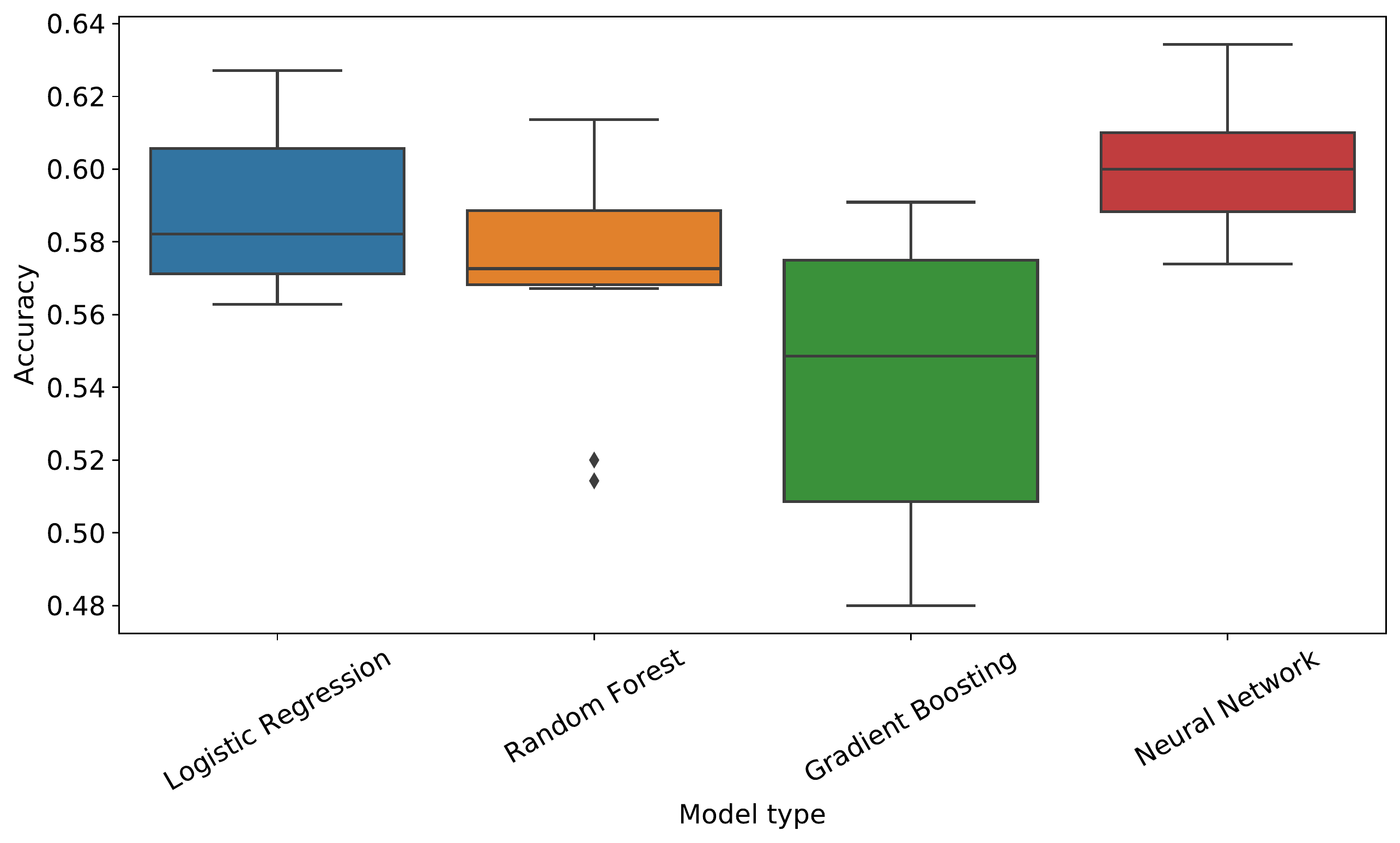}
\includegraphics[width=0.8\textwidth]{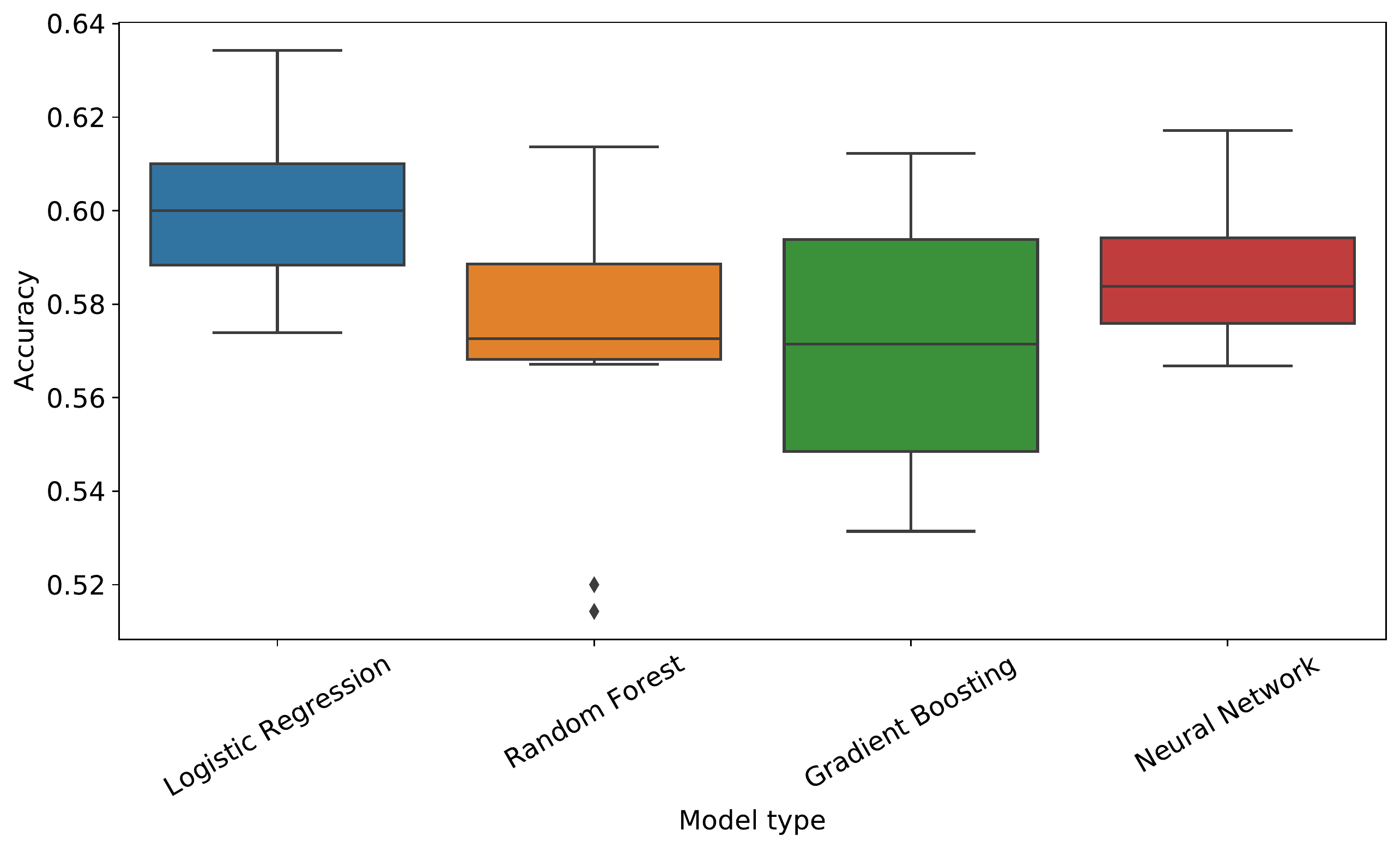}
\caption{A comparison of accuracy for different machine learning models: transferred models comparison (top) and ensemble model comparison (bottom).}
\label{Model_comparison}
\end{figure*}

Finally, we also tested the effect of using different machine learning classification algorithms for training the transferred models as well as for training the ensemble model (see Figure \ref{Model_comparison}).
The box plot on the top of the figure shows the effect of using different classifiers for training the transferred models (while keeping the ensemble classifier fixed --- using Logistic Regression).
As can be seen from the figure, using a Neural Network classifier slightly outperformed the others, but this difference was not found to be statistically significant.
More specifically, when applying an ANOVA test with a confidence level of 95\%, we rejected the null-hypothesis that all models performed the same.
Further applying the Tukey post hoc test, we found that Logistic Regression, Random Forest and Neural Network performed significantly better than Gradient Boosting, but the difference between the three models was not found to be significant.

The box plot on the bottom shows the effect of using different classifiers for training the ensemble model (while keeping the classifier of the transferred models fixed --- using Neural Network).
As can be seen from figure, using a Logistic Regression classifier slightly outperformed the others, but this difference was not found to be statistically significant.
More specifically, when applying an ANOVA test with a confidence level of 95\%, we could not reject the null-hypothesis that all models performed the same.

\section{Summary and Future Work}\label{conclusions}
In this paper, we suggested the Ensembled Transferred Embeddings learning framework.
The proposed framework is composed of four main steps:
1) manually label a small sample dataset;
2) extract embeddings from related large-scale labeled datasets;
3) train transferred models using the extracted transferred embeddings and the labels of the sample dataset; and
4) build an ensemble to combine the outputs of the different transferred models into a single prediction.
We then showed the applicability of the proposed framework for the item categorization task in settings for which the textual attributes representing items are noisy and short, and labels are not available.

We evaluated our method using a large-scale real-world invoice dataset provided to us by PayPal.
The results of this evaluation showed that our method significantly outperforms other traditional (e.g., TF-IDF) as well as state-of-the-art (e.g., methods based on general purpose pretrained models such as BERT) item categorization methods.
We believe that the reason for the superiority of our method is due to the unique characteristics of the text used in our setting --- item descriptions are short, noisy and are domain specific.
These characteristics make the text distribution in our setting different than that of the text used to train general purpose embeddings.
Consequently, this makes our embeddings more relevant for the task at hand, despite being trained with a much smaller dataset.

One limitation of our method is the need to manually label a high quality sample dataset.
As we explained in the paper, generating a high quality dataset, even when using a crowdsourcing service, is not a trivial task.
A non-negligible effort should be invested in properly designing the crowdsourced labeling task, in order to obtain an accurate result, in a relatively fast and cheap process.

Another limitation of our method is the use of domain knowledge.
In our framework domain knowledge is essential for the selection of the related large-scale labeled datasets.
As demonstrated throughout the paper, choosing datasets that are related to the domain and the task at hand is essential to building more accurate models.

\bigskip\noindent
In future work it would be interesting to investigate the following directions:
\begin{itemize}

\item \textbf{Uninformative items detection.} 15\% of the items that were annotated using Mechanical Turk were labeled as uninformative. A production item categorization system would need to filter out such items as a first step.
An interesting research direction would be to build a machine learning model to identify such uninformative items, perhaps using the same ETE framework.

\item \textbf{Using additional item attributes.} In this work we used only the textual attributes: item name and item description, as input for our model. Using other attributes such as price, seller ID, and user ID was shown in \citep{krishnan2019large} to improve the accuracy of the item categorization model.

\item \textbf{Hierarchical categories.} 
In this paper, we considered a confusion between fashion and baby-products the same as a confusion between electronics and food-n-drink, while cleverly the former is more tolerable.
E-commerce item categories can usually be described using a taxonomy tree, and such a hierarchy can be used to weight confusions differently.
For example, a confusion between two categories sharing a common ancestor at a deeper level of the tree should be weighted lower than a confusion between two categories sharing the root of the tree only.

\item \textbf{Additional domains and tasks.} We believe that the proposed ETE framework is generic enough and can be very useful in additional text classification domains such as health, law, social media, etc. We also believe that the ETE framework, with some adjustments, can be beneficial to tasks other than text classification, such as computer vision, audio processing, time series analysis, and more.

\item \textbf{Combining more methods for text classification with small data.} In recent years methods such as active learning and semi supervised learning showed great success in reducing the number of labeled samples needed for an accurate text classification. These methods can be integrated into the ETE framework in order to achieve better accuracy.

\end{itemize}

\section*{Acknowledgements}
\noindent
This research was funded by PayPal.
We would like to thank our colleagues from PayPal: Yaeli, Adam, Omer, and Avihay who provided meaningful insights and greatly assisted in improving this work.

\section*{References}
\bibliographystyle{model5-names}
\biboptions{authoryear}
\bibliography{refs}

\section{Appendix}
\label{Appendix}


\subsection{Number of terms used in the TF-IDF method}
\label{sec:tfidf}

\begin{figure}[H]
\centering
\includegraphics[width=0.9\linewidth]{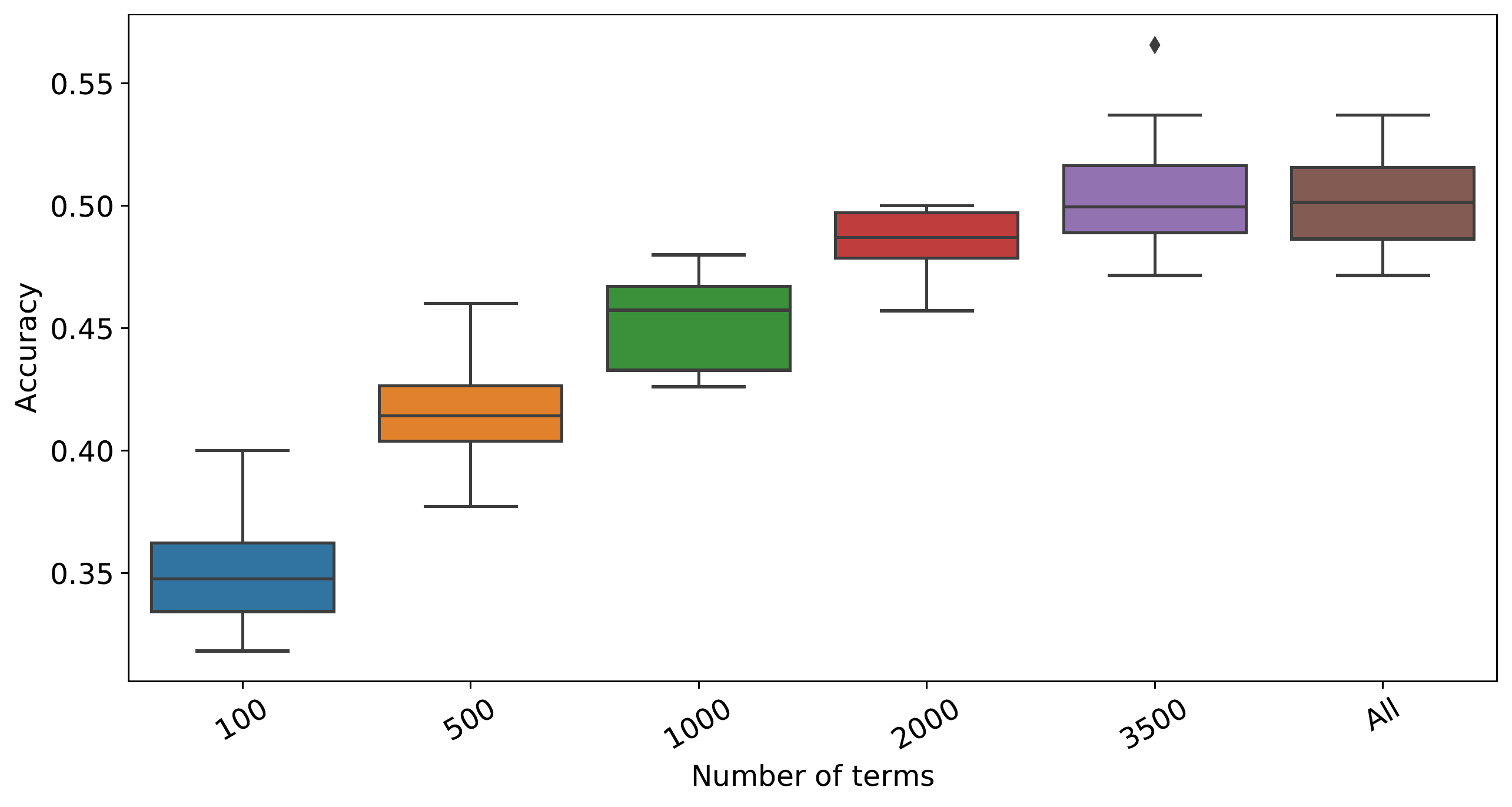}
\caption{A comparison of number of features for thr TF-IDF based model.}
\label{tfidf_terms}
\end{figure}

To determine the number of terms used in the TF-IDF method, we performed a simple analysis in which we executed the method with various numbers of terms, and calculated the average accuracy of the method over the different folds.
Figure \ref{tfidf_terms} presents the results of this analysis.
As can be seen from the figure, the method reaches the maximum accuracy when using approximately 3500 terms.

\subsection{Partial ensembles}
\label{sec:partial-ensembles}

\begin{figure}[H]
\centering
\includegraphics[width=0.9\linewidth]{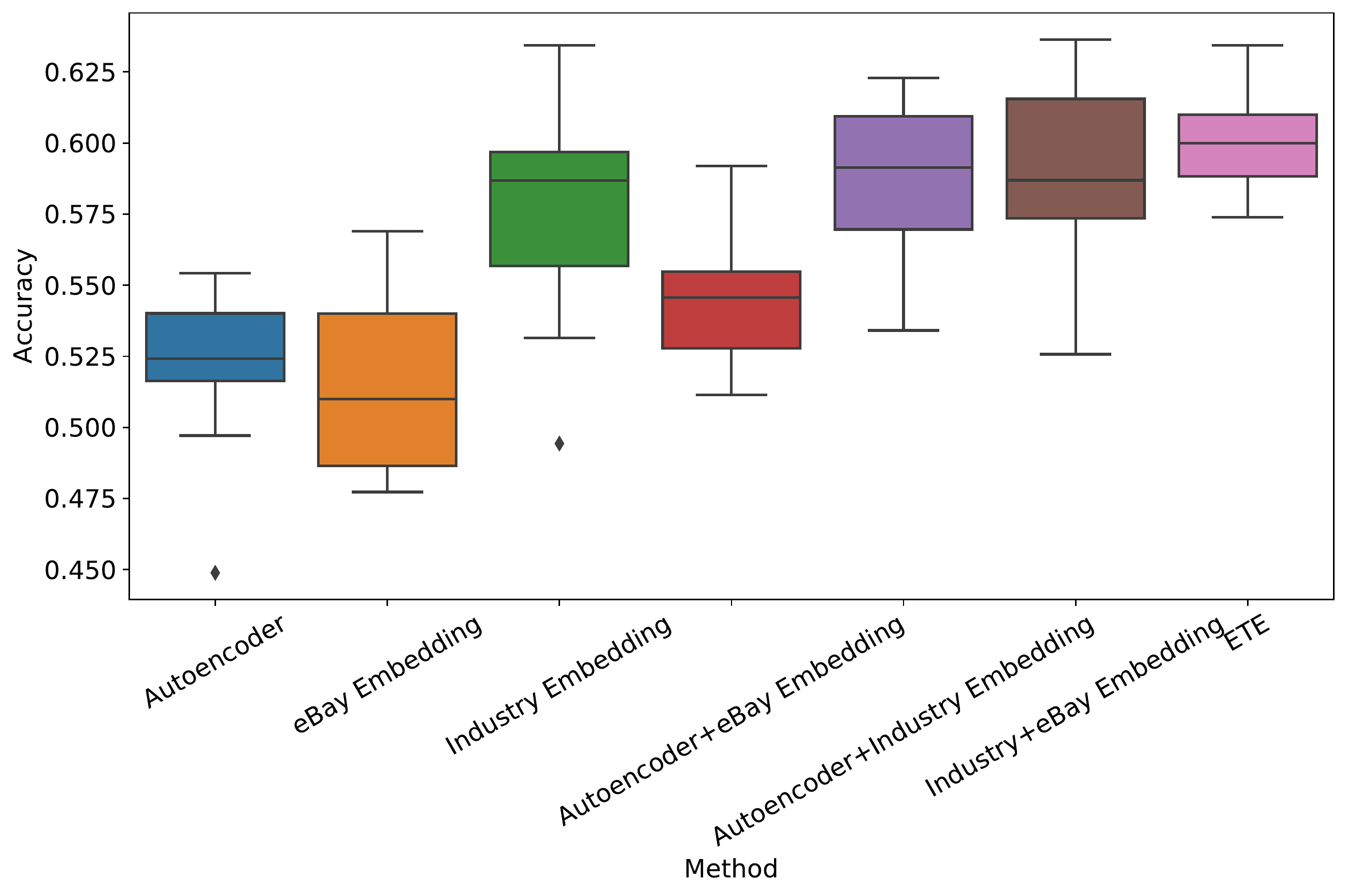}
\caption{A comparison of accuracy for all the combinations of partial embeddings in the ETE model.}
\label{partial_acc}
\end{figure}

Figure \ref{partial_acc} presents a comparison of ensembles based on all combinations of the Autoencoder, Industry Embedding and eBay Embedding methods.
As can be seen from the figure, the ETE method which ensembles all three embedding methods, presents the highest median, and the lowest interquartile range.
However, it is also apparent that most of classification power comes from the Industry Embedding method.

\subsection{Joining rare categories together}

\begin{figure}[H]
\centering
\includegraphics[width=0.9\linewidth]{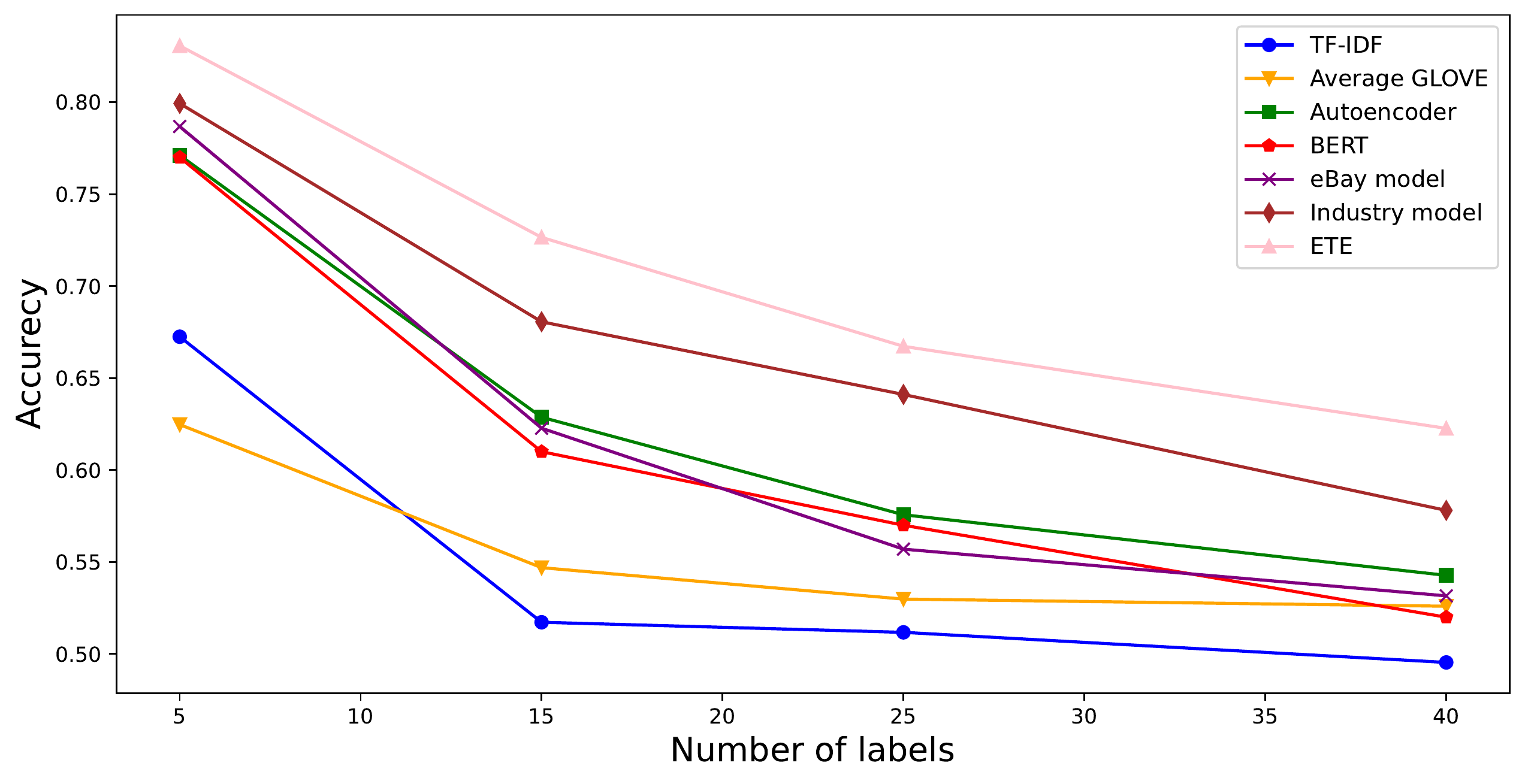}
\caption{A comparison of accuracy by the length of the item text.}
\label{num_categories_with_other}
\end{figure}

Figure \ref{num_categories_with_other} presents a similar analysis to the one presented in Figure \ref{Num_categories}, but now the rest of the categories are not ignored but rather joined into a single ``other'' category.
As can be seen, the two figures present quite similar trends.
Nevertheless, as expected the overall accuracy results in Figure \ref{Num_categories} are a bit higher.

\subsection{Error analysis}
\label{sec:error-analysis}

\begin{figure}[H]
\centering
\includegraphics[width=0.9\linewidth]{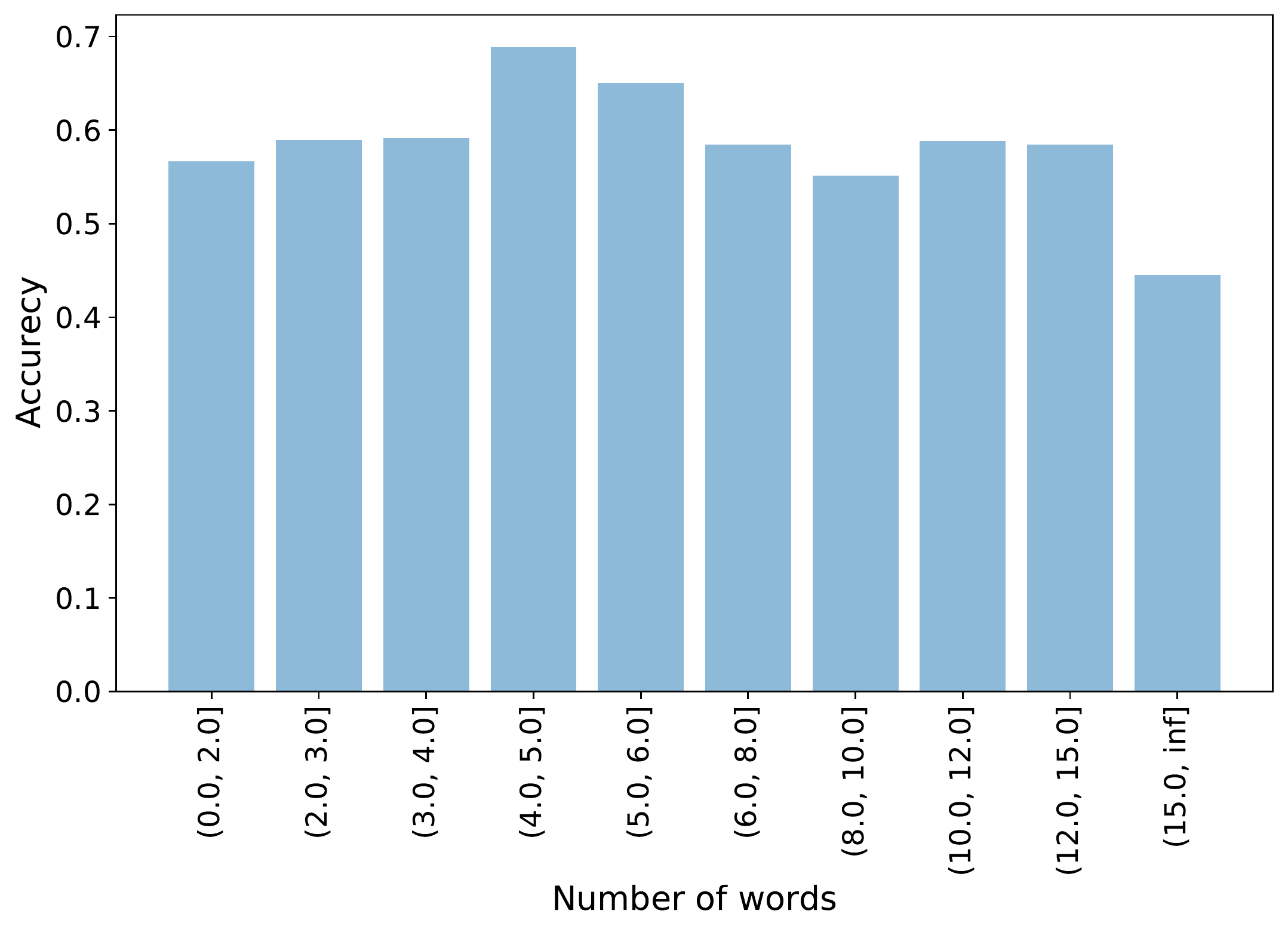}
\caption{A comparison of accuracy by the length of the item text.}
\label{len_comp}
\end{figure}

Figure \ref{len_comp} presents the accuracy of the ETE model on different lengths (words count) of item descriptions.
Examining the figure, it is difficult to identify a clear trend between the length of item description and accuracy.
Nevertheless, we do observe a reduction in accuracy for item descriptions with more than 15 words.
This makes sense since we used a maximum length of 15 tokens as input to our model (Note that this is not a real issue since only less than 5\% of the item descriptions have more than 15 words).

Table \ref{tab:recall_improvments} presents the relative improvement of the ETE model compared to the industry model (which seems to be our strongest baseline) for the 15 most popular categories.
As can be seen, categories such as sports-equip, baby-products and arts-n-craft benefit present the highest improvements in accuracy, most probably due to the incorporation of the eBay dataset in ETE.
In contrast, service categories such as services-other and website-services that do not exists in the eBay dataset show negative improvements in accuracy (i.e., the industry model performs better than the ETE model).

\begin{table}[h!]
\begin{center}
\caption{Relative improvement of the ETE model compared to the Industry model, measured by: F1 score, precision and recall. For convenience, we present the 15 most popular categories, sorted according to the relative improvement in F1 score.}
\label{tab:recall_improvments}
\resizebox{\textwidth}{!}{
\begin{tabular}{|c|c|c|c|} 
\hline
\textbf{Category}  & \textbf{F1 improvement} & \textbf{Precision improvement} & \textbf{Recall improvement} \\
\hline
baby-products    &    43.84\% &     41.55\% &  45.44\% \\
\hline
sports-equip     &    41.18\% &     42.71\% &  39.98\% \\
\hline
houseware        &    40.76\% &     35.00\% &  46.14\% \\
\hline
health           &    31.16\% &     33.86\% &  28.57\% \\
\hline
arts-n-craft     &    24.14\% &     21.07\% &  27.14\% \\
\hline
electronics      &    16.66\% &     13.04\% &  19.99\% \\
\hline
food-n-drink     &     7.95\% &      7.95\% &   7.95\% \\
\hline
auto-parts       &     2.90\% &      3.94\% &   1.75\% \\
\hline
jewelry          &     2.69\% &      5.22\% &   0.00\% \\
\hline
cosmetics        &    -2.11\% &     -8.88\% &   4.54\% \\
\hline
books            &    -7.77\% &      9.65\% & -28.57\% \\
\hline
books            &    -7.77\% &      9.65\% & -28.57\% \\
\hline
services-other   &   -15.05\% &    -19.31\% & -10.20\% \\
\hline
website-services &   -19.42\% &    -26.56\% & -12.50\% \\
\hline
\end{tabular}
}
\end{center}
\end{table}

\end{document}